\documentclass[10pt,twocolumn]{IEEEtran}%
\usepackage{verbatim}
\usepackage{subfigure}
\usepackage{float}
\usepackage{amsmath}
\usepackage{graphicx}
\usepackage{amssymb}
\usepackage{amsfonts}
\usepackage{psfrag}
\usepackage{subfigure}
\usepackage{algorithm}
\usepackage{algorithmic}
\usepackage{times}
\usepackage{fullpage}
\usepackage{placeins}
\usepackage{epstopdf}%
\providecommand{\U}[1]{\protect\rule{.1in}{.1in}}

\graphicspath{{figures//}}
\makeatletter
\floatstyle {ruled} \newfloat {algorithm}{tbh}{loa} \floatname {algorithm}{Algorithm}

\begin{document}

\title{Image Segmentation via Probabilistic Graph Matching}
\author{Ayelet Heimowitz\thanks{{Faculty of Engineering, Bar Ilan University,
Israel{}}{}. ayeltg@gmail.com.}, Yosi Keller\thanks{{Faculty of Engineering,
Bar Ilan University, Israel{}}{}. yosi.keller@gmail.com.}\\Faculty of Engineering, Bar-Ilan University, Israel.}
\date{}
\maketitle

\begin{abstract}
This work presents an unsupervised and semi-automatic image segmentation
approach where we formulate the segmentation as a inference problem based on
unary and pairwise assignment probabilities computed using low-level image
cues. The inference is solved via a probabilistic graph matching scheme, which
allows rigorous incorporation of low level image cues and automatic tuning of
parameters. The proposed scheme is experimentally shown to compare favorably
with contemporary semi-supervised and unsupervised image segmentation schemes,
when applied to contemporary state-of-the-art image sets.

\end{abstract}

\section{Introduction}

\label{sec:introduction}

Image segmentation is a fundamental problem in computer vision, and numerous
approaches for the efficient partitioning of images into several regions, or
subsets, were derived. In this work we study foreground-background
segmentation, where given an input image $I$ we aim to classify its pixels as
one of two mutually exclusive classes, $\mathcal{F}$ and $\mathcal{B}$,
corresponding to foreground and background image objects, respectively. We
consider the semi-automatic segmentation, where certain a-priori knowledge
concerning the sought after segmentation is provided. The a-priori cues are
given by associating certain regions or pixels within the processed image,
with either $\mathcal{F}$ or $\mathcal{B}$. Such a-priori information is
applicable in computer graphics applications (\textquotedblleft Magic
Wand\textquotedblright\ etc.), as well as in detection-based approaches
\cite{girshick14CVPR,BharathECCV2014,sliding_window,geometric}, that first
recover the foreground's bounding box.

Contemporary foreground-background ($\mathcal{F}$/$\mathcal{B}$) segmentation
schemes introduce unary and pairwise potentials \cite{RotherKB, BoykovJolly,
BoxPrior, starPrior, MalikShi, Topological}. A unary potential quantifies the
similarity of a pixel to the classes $\mathcal{F}$ and $\mathcal{B}$, while
the pairwise term quantifies the similarity between pairs of pixels. Pairwise
and unary terms can be represented via graphs, and the optimization of
functionals involving them, can be formulated via graph algorithms.

In their seminal work \cite{ShiNCI}, Shi and Malik formulated the image
segmentation problem as a graph partitioning problem, where the resulting
minimization is shown to be NP-hard, and the solution was thus approximated
via spectral relaxation. This approach is related to spectral embeddings and
clustering \cite{NIPS2001_2092} that allows multiclass clustering, and the
introduction of constraints into the spectral formulation \cite{ShiNCI, Xu,
Yu}.

Boykov and Jolly \cite{BoykovJolly} proposed to solve the min-cut problem by
the \textit{graph-cuts} approach that formulates the min-cut minimization as a
graph-based maximum flow problem, following the \textquotedblleft Max-flow
min-cut\textquotedblright\ theorem. This is a discrete optimization approach
that paved the way for a gamut of extensions in foreground-background
segmentation\ \cite{RotherKB, BoykovJolly, BoxPrior, starPrior, Topological},
that utilized the graph-cut solver and its extensions \cite{1238310}.

These approaches (as well as the proposed scheme)\ are \textit{unsupervised}
as the class of the segmented objects is unknown apriori and the unary
potentials are learnt based on the input image alone. Hence, such approaches
can be applied to images containing any classes of objects. In contrast,
\textit{supervised} schemes utilize a training set of images for object
segmentation \cite{sliding_window,long_shelhamer_fcn,Zhen_ICCV15_CRFRNN} and
detection \cite{renNIPS15fasterrcnn}. Segmentation schemes can also be either
\textit{automatic} or \textit{semi-automatic}, where automatic schemes
\cite{Zhen_ICCV15_CRFRNN} are only given the input image with no user
interaction, while semi-automatic schemes, such as GrabCut utilize user
interaction to coarsely localize the object of interest (foreground). With the
significant advance in Deep Learning-based object detection
\cite{renNIPS15fasterrcnn}, the need for manual coarse foreground localization
is less essential.

In this work, we formulate the foreground-background segmentation as a
semi-automatic unsupervised probabilistic inference problem, where we first
aim to estimate the marginal assignment probabilities of image elements
$\left\{  \mathbf{x}_{i}\right\}  $, superpixels in our implementation, to
either of the labels $L=\left\{  \mathcal{F},\mathcal{B}\right\}  $. The image
segmentation is thus given by their maximum likelihood assignments%

\begin{equation}
p\left(  \mathbf{x}_{i}\right)  =\left[  p\left(  \mathbf{x}_{i}\in
\mathcal{F}\right)  ,p\left(  \mathbf{x}_{i}\in\mathcal{B}\right)  \right]
^{T} \label{equ:marginal}%
\end{equation}
Our approach utilizes \textit{empirically estimated} pairwise assignment
probabilities
\begin{multline}
p\left(  \mathbf{x}_{i},\mathbf{x}_{j}\right)  =\label{equ:pairwise}\\
\left[  p\left(  \mathbf{x}_{i}\in\mathcal{F},\mathbf{x}_{j}\in\mathcal{F}%
\right)  ,...,p\left(  \mathbf{x}_{i}\in\mathcal{B},\mathbf{x}_{j}%
\in\mathcal{B}\right)  \right]  ^{T}%
\end{multline}
and unary assignment probabilities,%
\begin{equation}
p_{u}\left(  \mathbf{x}_{i}\right)  =\left[  p\left(  \mathbf{x}_{i}%
\in\mathcal{F}\right)  ,p\left(  \mathbf{x}_{i}\in\mathcal{B}\right)  \right]
^{T}, \label{equ:unary}%
\end{equation}
respectively.

The inference is derived using the Probabilistic graph matching (PGM) approach
introduced by Egozi et al. \cite{KellerProbSpec}, where it was shown that
spectral graph matching can be applied to inference problems. The PGM approach
introduces two inference schemes. The first applies spectral relaxation to a
NP-hard inference problem, similar to Spectral min-cut approaches
\cite{ShiNCI,ShiNCI, Xu, Yu}, and the second is an iterative Bayesian
inference scheme, providing improved inference accuracy. In contrast to both
spectral and Max-flow approaches, no graph-based metric (such as
min-cut/max-flow) is optimized by the PGM, and the result is the set of
marginal assignment probabilities $p\left(  \mathbf{x}_{i}\in\mathcal{L}%
\right)  $, that allow maximum likelihood image segmentation. The iterative
Bayesian inference \cite{KellerProbSpec} utilizes the estimated marginal
assignment $p_{m}\left(  s_{i}\in L_{k}\right)  $ probabilities to adaptively
reweigh the pairwise assignment probabilities $p\left(  s_{i}\in L_{k_{1}%
},s_{j}\in L_{k_{2}}\right)  $. Thus, for $p_{m}\left(  s_{1}\in B\right)
\approx0$, we set $p\left(  s_{1}\in B,s_{j}\in L_{k_{2}}\right)  =0$ $\forall
j$, and the inference can be recomputed using the refined pairwise probabilities.

As the classification of each pixel individually might prove exhaustive in
terms of memory and computational power, we follow recent approaches
\cite{MemoryReduction} that utilize superpixels\textit{ }(SPs) by partitioning
the input image into small homogeneous regions. In this work we apply our
approach using color-based binary and unary assignment probabilities,
following the potentials used in previous works \cite{RotherKB, BoykovJolly,
BoxPrior, starPrior, Topological}. Yet, it can be applied with any viable
assignment probabilities, such as those derived by Deep Learning
\cite{long_shelhamer_fcn,Zhen_ICCV15_CRFRNN}.

As such we present the following contributions:

\textbf{First}, we propose a probabilistic inference approach to image
segmentation, that is solved using probabilistic graph matching.
\textbf{Second}, we show how to extend the PGM approach to utilize unary
probabilities, that were not used in the original formulation by Egozi et al.
\cite{KellerProbSpec}. \textbf{Third}, the scheme is shown to compute a
maximum-likelihood estimate of the segmentation, in contrast to prior works,
that optimized graph-based measures such as minimal-cut and max-flow.
\textbf{Last}, we propose an iterative refinement scheme for both the
segmentation (as in prior schemes), and the parameters used to compute the
assignment probabilities. We present promising results for both semi-automatic
and automatic segmentations.

This paper is organized as follows: In Section \ref{sec:background} we review
prior approaches to $\mathcal{F}$/$\mathcal{B}$ image segmentation. We detail
the proposed approach in Section \ref{sec:Spectral Grabcut}, and report the
experimental results in Section \ref{sec:experiemental}, where we compare
against contemporary state-of-the-art schemes using standard sets of test
images. Concluding remarks are presented in Section \ref{sec:discussion}.

\section{Background}

\label{sec:background}

Shi and Malik formulated image segmentation as a graph partitioning problem
\cite{ShiNCI}, solved by minimizing the normalized cuts objective function via
spectral relaxation, resulting in a generalized eigenvalue problem. Their
seminal work paved the way to multiple extensions\textbf{.} Yu and Shi studied
constrained image segmentation into $K$ disjoined partitions \cite{Yu}, by
applying the normalized cuts criterion, and a relaxing the problem, utilizing
the $K$ leading eigenvectors of the corresponding affinity matrix. Xu et al.
optimized a relaxed formulation of the normalized cuts criterion under linear
constraints \cite{Xu}, as an eigendecomposition, solved by the projection of
the leading eigenvector onto the subspace of linear constraints.\textbf{ }The
segmentation problem was formulated as a spectral clustering problem by Ng et
al \cite{NIPS2001_2092}. The $K$ leading eigenvectors of the normalized affinity matrix
are used as the embedding of the image pixels in $R^{K}$. These points can be
clustered into $K$ distinct clusters using K-Means.

In the seminal work of Boykov and Jolly \cite{BoykovJolly}, the user provides
hard segmentation constraints by labeling certain pixels as \textquotedblleft
object\textquotedblright\ and other as \textquotedblleft
background\textquotedblright. These are denoted as \textit{hard constraints,
}as the resulting segmentation has to adhere with this labeling. The labeled
pixels allow to compute intensity distribution models for the object and
background. Soft segmentation constraints are provided through a cost
function, which considers the region and boundary properties of the segment,
and the segmentation is derived by Graph-Cuts minimization of a binary cost function.

Rother \emph{et al}. extended the graph-cut approach by the GrabCut framework
\cite{RotherKB}, where the user provides as input an initial TriMap
$T=\{T_{B},T_{U},T_{F}\}$. The background region, $T_{B}$, is specified by the
user, such that a pixel $p\in T_{B}$ has to be labeled as background. The
$\mathcal{F}$ and $\mathcal{B}$ regions are modeled by Gaussian mixture models
(GMMs), that define unary potentials, that are iteratively optimized alongside
pairwise terms using Graph Cuts.

Lempitsky \emph{et al}. \cite{BoxPrior} proposed to improve the accuracy of
the initial $\mathcal{F}$/$\mathcal{B}$ models, and consequently the
segmentation accuracy of \cite{RotherKB,BoxPrior} by first computing an
initial background model, and using a third of the pixels within the bounding
box, most different from the background model, to train the initial foreground
model. They also propose a tightness prior assuming that the bounding box is
adjacent to the edges of the object, such that the prior is combined with the
Graph-Cut cost function.

Shape priors allow to restrict the segmentation results to a particular class
of shapes. Veksler \emph{et al}. \cite{starPrior} propose a star-based shape
model, that is combined with terms that encode the region and boundary
properties to derive a modified cost function. As Graph-Cuts schemes show a
bias toward small segments, Veksler proposes a prior term that encourages
larger segments. The compact shape prior introduced by Das \emph{et al}.
\cite{CompactPrior} is combined with the graph-cut framework, and allows to
reduce the required user interaction. The authors also propose a
\textquotedblleft quality check\textquotedblright\ measure of the resulting
segmentation, that allows to automatically refine the schemes parameters.

The use of shape prior for medical imaging was studied by Freedman \emph{et
al}. \cite{MedicalShapePrior}, as such images often contain multiple objects
lacking salient edges, that might cause graph-cut-based approaches to fail. It
was shown that the use of shape priors in the graph-cut framework resolves the
problem. Slabaugh \emph{et al}. utilize an elliptical shape prior
\cite{EllipticalShapePrior} that is shown to be efficient in the segmentation
of blood vessels and lymph nodes. Chen \emph{et al}. derived a novel approach
for inducing topological constraints \cite{Topological}, by defining an energy
function that combines unary and pairwise potentials, which is minimized
subject to topological {considerations}.

Lazy snapping \cite{LazySnapping} is a semi-automatic image segmentation
approach related to the graph-cut framework, which utilizes additional user
interaction to improve upon graph-cut segmentation. The nodes of the graph are
SPs, reducing the computational complexity, and provides instant visual
feedback. A secondary marking scheme is employed to further improve the
segmentation. Krahenbuhl and Koltun \cite{NIPS2011_4296} formulated the image
segmentation as an inference over a fully connected Conditional Random Field
(CRF) model, that is constructed over pixels rather than SPs, using a
computationally efficient approximate inference scheme based on a linear
combination of Gaussian kernels.

The work of Alpert et al. \cite{Alpert2012} is of particular interest to us,
as it presents a probabilistic multiscale bottom-up approach to image
segmentation. In particular, the authors show how to quantify \textit{unary}
assignment probabilities based on low level image cues such as color and
texture. These unary cues are fused using a \textquotedblleft Mixture of
experts\textquotedblright\ formulation. A geometric prior is used to encode
the geometry of the regions. A coarse-to-fine approach propagates the
classification probabilities from the coarse resolution scales to the finer
ones. In contrast, the focal point of our scheme is the solution of
probabilistic high order assignment program that utilizes both unary and
pairwise cues.

Carreira et al. proposed the Constrained Parametric Min-Cuts (CPMC) approach
\cite{CPMC}, that starts by detecting multiple segmentations by computing
multiple binary min-cuts segmentations in multiple image resolutions. The
resulting set of segments is pruned by removing trivial solutions and applying
a ranking regressor trained to predict the validity of the resulting segments.

A segmentation prior for estimating the foreground's spatial support was
introduced by Rosenfeld and Weinshall \cite{geometric}. It is a
general-purpose object detector, which provides a coarse localization of the
foreground in an image. For each test image the GIST descriptor \cite{Oliva}
is computed, and the most similar training images are found. A summation of
the ground truth classification of these images is used to approximate the
foreground classification probability in the test image. {A supervised}
approach for learning the unary affinities was suggested by Kuettel and
Ferrari \cite{sliding_window}, by partitioning the input image into multiple
foreground windows. These are matched to their nearest neighbors in a set of
training windows, marked in images annotated with foreground-background masks.
Thus, deriving the unary potentials of a binary objective function, that is
minimized via graph-cuts.

Hariharan et al. proposed a Simultaneous Detection and Segmentation
\cite{BharathECCV2014} approach, that utilizes state-of-the-art object
detection schemes \cite{girshick14CVPR}, based on Deep-Learning. The detection
is used to initiate the segmentation scheme by providing a bounding box
hypothesis. In particular, such a scheme is \textit{category-specific} as it
is based on a category-specific object detector and learning set. In contrast,
prior works, as well as ours, do not assume any particular object category or
learning set.

Image partitioning using superpixels \cite{WatershedB, turbo} is commonly used
in segmentation schemes. In our work we used the watershed algorithm
\cite{WatershedB} as it provides accurate over-segmentation without geometric
regularization that might hamper the over-segmentation accuracy.

\section{Image segmentation by probabilistic graph matching}

\label{sec:Spectral Grabcut}

In Foreground/Background ($\mathcal{F}$/$\mathcal{B)}$ image segmentation,
each image element $s_{i}$ (pixels, SPs) is classified as either $\mathcal{F}$
or $\mathcal{B}$, such that $s_{i}\in\mathcal{F}$ or $s_{i}\in\mathcal{B}$.
Let $S=\{s_{i}\}_{i=1}^{n}$ be the set of SPs in an input image $I$. The core
of our approach is to estimate the marginal assignment probabilities
\begin{equation}
p\left(  \mathbf{x}_{i}\right)  =\left[  p\left(  \mathbf{x}_{i}\in
\mathcal{F}\right)  ,p\left(  \mathbf{x}_{i}\in\mathcal{B}\right)  \right]
^{T},
\end{equation}
based on the empirical unary and pairwise assignment probabilities as in Eqs.
\ref{equ:unary} and \ref{equ:pairwise}, that are detailed in Sections
\ref{subsec:gmm} and \ref{subsec:pairwise}, respectively.

Our approach is fully unsupervised, based on the input image \textit{only},
and thus differs form recent works on Deep Learning based semantic
segmentation \cite{long_shelhamer_fcn}, that require a large training set
consisting of the classes of objects that might appear in the image. However,
given a class-specific detector \cite{girshick14CVPR}, or a training set
\cite{sliding_window}, their output can be incorporated into the proposed
schemes, as either priors or unary and pairwise probability models.

A gamut of image cues such as color, texture, and contours, to name a few,
were used in previous works \cite{5204091}. These can be encoded by different
image descriptors (SIFT, HOG, GIST, etc.) and representations (histograms,
GMMs, Gaussians, etc.). A segmentation scheme can utilize multiple cues and
representations \cite{Alpert2012}. In this work, we use color as a single cue
for simplicity and comparison to prior schemes in $\mathcal{F}$/$\mathcal{B}$\ segmentation.

\subsection{Unary assignment probabilities}

\label{subsec:gmm}

The unary probabilities $\left[  p_{u}\left(  \mathbf{s}_{i}\in\mathcal{F}%
\right)  ,p_{u}\left(  \mathbf{s}_{i}\in\mathcal{B}\right)  \right]  ^{T}$ of
an SP $s_{i}$, encoded by a Gaussian model \textbf{$G_{i}$ }of the pixels in
the LAB color space, are estimated by their distance to the GMM color models
of the foreground and background, $GMM_{F}$ and $GMM_{B}$, respectively, using
a Radial Basis Functions (RBF) kernel%

\begin{equation}
p_{u}\left(  \mathbf{s}_{i}\in\mathcal{L}\right)  \propto\exp\left(
-\frac{\tilde{D}_{KL}(G_{i}\Vert GMM_{\mathcal{L}})}{\sigma_{u}}\right)
\newline\newline,
\end{equation}
where
\begin{equation}
\tilde{D}_{KL}(G_{i}\Vert GMM)=\min_{j}(D_{KL}(\mathbf{G}_{i}\Vert o_{j}%
)-\log(\alpha_{j})) \label{klGMM}%
\end{equation}
is an approximate KL divergence \cite{gold04}. The GMM model%

\begin{equation}
GMM=\mathbf{\Sigma}_{k=1}^{K}\alpha_{k}o_{k},
\end{equation}
consists of $K$ Gaussians, and $\alpha_{k}$ is the prior of a Gaussian $o_{k}%
$, and the RBF bandwidth $\sigma_{u}$ is set to%
\begin{equation}
\sigma_{u}=\underset{G_{i}\in\mathcal{F}}{median}\tilde{D}_{KL}(G_{i}\Vert
GMM_{\mathcal{F}}). \label{equ:unary sigma}%
\end{equation}
which is a robust (median-based) maximum-likelihood estimate of a Gaussian's bandwidth.

\subsection{Pairwise assignment probabilities}

\label{subsec:pairwise}

The pairwise assignment probabilities $p\left(  s_{i},s_{j}\right)  $ quantify
the probability of two SPs to have the same assignment, where similar SPs are
expected to have the same labels%
\begin{align}
p\left(  s_{i}\in\mathcal{F},s_{j}\in\mathcal{F}\right)   &  =p\left(
s_{i}\in\mathcal{B},s_{j}\in\mathcal{B}\right) \label{equ:rbf prob same}\\
&  \propto\exp\left(  -D_{KL}^{s}(G_{i}\Vert G_{j})/\sigma_{p}\right)
,\nonumber
\end{align}
where $D_{KL}^{s}(G_{i}\Vert G_{j})$, is a the symmetric Kullback-Liebler (KL)
Divergence \cite{gold04} between Gaussians%
\begin{equation}
D_{KL}^{s}(s_{i}\Vert s_{j})=\min(D_{KL}(s_{i}\Vert s_{j}),D_{KL}(s_{j}\Vert
s_{i})). \label{equ:sym gauss KL}%
\end{equation}
such that
\begin{multline}
2D_{KL}(G_{i}\Vert G_{j})=\log\left(  {\frac{|\mathbf{\Sigma}_{j}%
|}{|\mathbf{\Sigma}_{\mathbf{i}}|}}\right)  +\mathrm{tr}\left(  \mathbf{\Sigma
}_{j}^{-1}\mathbf{\Sigma}_{\mathbf{i}}\right) \label{equ:gauss KL}\\
+\left(  \mathbf{\mu}_{j}-\mathbf{\mu}_{i}\right)  ^{\top}\mathbf{\Sigma}%
_{j}^{-1}(\mathbf{\mu}_{j}-\mathbf{\mu}_{i}),
\end{multline}
where \textbf{$\Sigma$}$_{i}$ and \textbf{$\mu$}$_{i}$ are the covariance and
mean, respectively, of the Gaussian $G_{i}$. Equation \ref{equ:gauss KL} is
symmetrized as in Eq. \ref{equ:sym gauss KL} by computing the minimal KL
divergence, as the distance between a Gaussian with an ill-conditioned
covariance matrix \textbf{$\Sigma$}$_{i}$ to any other Gaussian might be
large, even for similar SPs. In that we follow the work of Goldberger et al.
\cite{gold04}, where the KL-divergence between two GMMs $GMM_{1}$ and
$GMM_{2}$ is approximated using the minimal KL-divergence between components
of $GMM_{1}$ and $GMM_{2}$.

As an SP\ can be assigned to either $\left\{  \mathcal{B},\mathcal{F}\right\}
$ we have that
\begin{align}
p\left(  s_{i}\in\mathcal{B},s_{j}\in\mathcal{F}\right)   &  =p\left(
s_{i}\in\mathcal{F},s_{j}\in\mathcal{B}\right) \label{equ:rbf prob notsame}\\
&  \propto1-p\left(  s_{i}\in\mathcal{B},s_{j}\in\mathcal{B}\right)
,\nonumber
\end{align}

where the RBF bandwidth $\sigma_{p}$ is given by%
\begin{equation}
\sigma_{p}=medianD_{KL}^{s}(G_{i}\Vert G_{j}),G_{i},G_{j}\mathbf{\in
\{\mathcal{F},\mathcal{B}\}} \label{equ:pair sigma}%
\end{equation}
\textbf{ }

As each SP can be assigned to a single class
\begin{multline}
p\left(  s_{i}\in\mathcal{B},s_{j}\in\mathcal{B}\right)  +p\left(  s_{i}%
\in\mathcal{F},s_{j}\in\mathcal{F}\right)  +\label{equ:pairwise normalization}%
\\
p\left(  s_{i}\in\mathcal{B},s_{j}\in\mathcal{F}\right)  +p\left(  s_{i}%
\in\mathcal{F},s_{j}\in\mathcal{B}\right)  =1.
\end{multline}

\subsection{Probabilistic inference}

\label{subsec:assignment}

Given the unary and pairwise assignment probabilities, estimated as in
Sections \ref{subsec:gmm} and \ref{subsec:pairwise}, we aim to derive an
inference scheme able to estimate the marginal assignment probabilities as in
Eq. \ref{equ:marginal}. Quadratic inference problems with discrete labels are
known to be NP-hard, as they can be formulated as second order Markov Random
Field (MRF) inference. Yet, efficient approximate solutions have been derived,
such as Graph Cuts \cite{Boykov2001} and Loopy Belief Propagation (LBP)
\cite{5539797,Weiss2001}.

In this work we apply the Probabilistic Graph Matching approach by Egozi et
al. \cite{KellerProbSpec}, who showed that spectral graph matching (SGM)
\cite{Leordeanu-spectral} can be applied to efficiently approximate MRF
inference, yielding a maximum likelihood estimate of the marginal assignment
probabilities $p\left(  \mathbf{s}_{i}\right)  $. The authors also proposed
the Probabilistic Graph Matching (PGM) improved inference scheme, based on
iterative Bayesian estimation. In contrast to Graph Cuts (Max Flow) schemes
that optimize graph-related metrics, SGM, PGM, and LBP\ estimate marginal
probabilities. Moreover, SGM and PGM can be applied to pairwise probabilities
represented by dense graphs \cite{KellerOfdm}, while the LBP is commonly
applied to sparse, tree-like graphs.

Both SGM and PGM are applied to the matrix of pairwise assignment
probabilities $\mathbf{P\in%
\mathbb{R}
}_{+}^{2n\times2n}$, such that%
\begin{align}
p_{2i-1,2j-1}  &  =p\left(  s_{i}\in\mathcal{\mathbf{F}},s_{j}\in
\mathcal{\mathbf{F}}\right) \\
p_{2i,2j}  &  =p\left(  s_{i}\in\mathcal{\mathbf{B}},s_{j}\in
\mathcal{\mathbf{B}}\right) \nonumber\\
p_{2i-1,2j}  &  =p_{2i,2j-1}=p\left(  s_{i}\in\mathcal{B},s_{j}\in
\mathcal{F}\right) \nonumber
\end{align}
and normalized according to Eq. \ref{equ:pairwise normalization}, such that%
\[
p_{2i-1,2j-1}+p_{2i,2j}+p_{2i-1,2j}+p_{2i,2j-1}=1.
\]

\subsubsection{Integrating pairwise and unary probabilities}

The probabilistic inference formulation discussed by Egozi et al.
\cite{KellerProbSpec}, only utilized pairwise assignment probabilities, as it
was derived from SGM. In order to incorporate the estimated unary
probabilities $p_{u}\left(  \mathbf{s}_{i}\right)  $, we note that one can
recover the marginals $p\left(  \mathbf{s}_{i}\right)  $ as the leading
eigenvector of the diagonal matrix $\mathbf{C\in%
\mathbb{R}
}_{+}^{2n\times2n}$ such that
\[
c_{_{2i-1,2i-1}}=p_{u}\left(  s_{i}\in\mathcal{\mathbf{F}}\right)
^{2},c_{_{2i,2i}}=p_{u}\left(  s_{i}\in\mathcal{\mathbf{B}}\right)  ^{2}%
\]

Thus, the marginals $p\left(  \mathbf{s}_{i}\right)  $ can be estimated by
applying either the SGM or PGM inference schemes to $\overline{\mathbf{P}%
}\mathbf{\in%
\mathbb{R}
}_{+}^{2n\times2n}$%
\begin{equation}
\overline{\mathbf{P}}=\mathbf{P+\lambda}^{2}\mathbf{C,} \label{equ:lamda}%
\end{equation}
where $\mathbf{\lambda}^{2}$ is a weighting factor set manually, that balances
the terms. $\overline{\mathbf{P}}$ is symmetric with non-negative entries, and
it follows by the Perron-Frobenius theorem that it is guaranteed to have a
single eigenvector (at least) with non-negative entries. This property
guarantees the numerical robustness of the SGM and PGM schemes.

\subsection{Iterative refinement and parameters auto-tuning}

\label{subsec:refinement}

The inference scheme proposed in Section \ref{subsec:assignment} can be
extended by iteratively refining the segmentation and GMM representation of
$\mathcal{F}$ and $\mathcal{B}$, by training them on the refined sets of SPs
$\mathcal{F}_{t}$ and $\mathcal{B}_{t}$. In turn, this allows to refine the
estimation of the unary and pairwise assignment probabilities.

Let $\mathcal{F}_{t}=\left\{  s_{i}\in\mathcal{F}\right\}  _{t}$ and
$\mathcal{B}_{t}=\left\{  s_{i}\in\mathcal{B}\right\}  _{t}$ be the sets of
SPs assigned to $\mathcal{F}$ and $\mathcal{B}$ in iteration $t$,
respectively. Following the probabilistic interpretation in Section
\ref{subsec:assignment}, we aim to derive a maximum likelihood auto-tuning
scheme for the bandwidth of the RBF kernels used to estimate the assignment
probabilities. The core of our auto-tuning approach is to estimate the
$\sigma_{u}^{B}$ and $\sigma_{u}^{F}$, the RBF bandwidths, as the average
distance within each class, corresponding to a maximum likelihood estimate of
the parameters of a Gaussian model%
\begin{equation}
\sigma_{u}^{F}=\frac{1}{\left\vert \mathcal{F}_{t}\right\vert }%
{\displaystyle\sum\limits_{s_{i}\in\mathcal{B}_{t}}}
KL\left(  GMM_{F}^{t},s_{i}\right)  \label{equ:sigmaub}%
\end{equation}
and $\sigma_{u}^{B}$ is computed mutatis mutandis\textbf{. }

As for the pairwise bandwidths, we computed three bandwidths, $\sigma_{p}^{B}%
$, $\sigma_{p}^{F}$, $\sigma_{p}^{B/F}$ corresponding to the distances within
the classes $KL\left(  s_{i}\in\mathcal{B},s_{j}\in\mathcal{B}\right)  $,
$KL\left(  s_{i}\in\mathcal{F},s_{j}\in\mathcal{F}\right)  $, and the
inter-class distance $KL\left(  s_{i}\in\mathcal{F},s_{j}\in\mathcal{B}%
\right)  $. The pairwise bandwidths are computed by%
\begin{equation}
\sigma_{p}^{B}=\frac{1}{\left\vert \mathcal{B}_{t}\right\vert }%
{\displaystyle\sum\limits_{s_{i},s_{i}\in\mathcal{B}_{t}}}
KL\left(  s_{i},s_{j}\right)  , \label{equ:sigmapb}%
\end{equation}
where the other bandwidths are computed mutatis mutandis. The distance between
the object model and the background model is estimated as the median distance
between SPs classified as $\mathcal{F}$ and $\mathcal{B}$ SPs in the latest iteration.

\subsection{Extension to fully automatic segmentation}

\label{subsection:Unsupervised}

In order to extend the proposed scheme to image segmentation with no user
input, an object detection scheme is required to provide an initial estimate
of $\mathcal{F}$/$\mathcal{B}$. There are a gamut of object detection schemes
\cite{geometric,girshick14CVPR}, where the most accurate are supervised class
specific \cite{girshick14CVPR}, namely, meant to detect a particular class of
objects. As such schemes are supervised, one can learn additional cues, such
as texture, color and shape, that can be used to improve the initial estimate
of the $\mathcal{F}$/$\mathcal{B}$ models.

In this work, in sake of simplicity, aiming to retain the focal point of the
proposed scheme, and being able to directly compare against prior results, we
implemented the prior proposed by Rosenfeld and Weinshall \cite{geometric},
that coarsely estimates the general location of the foreground in an image, by
computing the foreground probability. We set a high background detection
threshold, such that we are assured that the initial estimate of the
foreground encloses the actual foreground object.

\subsection{Implementation issues and future extensions}

\label{subsec:implementation}

We tested various approaches for estimating the optimal rank of the GMM models
based on the AIC and BIC criteria \cite{GMM-BIC,MCLA2000}, but these did not
prove efficient. As both the foreground and background might be comprised of
multiple visually dissimilar regions, dissimilar neighboring SPs might belong
to the same object. Hence, local \textit{dissimilarity} of SPs is less
meaningful than local \textit{similarity}, as similar neighboring SPs always
relate to the same label.

The proposed scheme can be extended in future in several ways. First, the
representation of SPs and $\mathcal{F}$/$\mathcal{B}$ via Gaussians and GMMs
respectively, entails numerical difficulties. First, setting of the number GMM
components, and second, the numerical instability of the representation of
smooth image patches, having singular covariance matrices. Thus, causing the
computation of the $KL$ distance between SPs to be numerically unstable and
inaccurate. For that, we propose to utilize a histogram-based approach for SP
and object representation. Such a representation would be based on computing
an adaptive dictionary for the $\mathcal{F}$/$\mathcal{B}$ objects and SPs.

Second, the use of the probabilistic framework allows to pave the way for a
multiscale Bayesian formulation where the assignment probabilities inferred in
a coarse scale are used as priors in the succeeding finer resolution scale.
Thus, Eq. \ref{equ:rbf prob same} can be reformulated as
\begin{equation}
p\left(  \left(  s_{i},s_{j}\right)  \in\mathcal{F}|s_{i}\in\mathcal{F}%
,s_{j}\in\mathcal{F}\right)  \propto\exp(-\frac{D_{KL}^{s}(G_{i}\Vert G_{j}%
)}{\sigma_{p}}), \label{equ:refined}%
\end{equation}
where $p\left(  \left(  s_{i},s_{j}\right)  \in\mathcal{F}\right)  $ is the
pairwise probability that \textit{both} $s_{i}$ and $s_{j}$ are labeled as
$\mathcal{F}$, while the term $s_{i}\in\mathcal{F}$ relates to the unary assignment.

Note that in the formulation presented in Section \ref{subsec:pairwise} the
pairwise and unary probabilities are derived from different image cues. Thus,
$p\left(  \left(  s_{i},s_{j}\right)  \in\mathcal{F}\right)  =p\left(  \left(
s_{i},s_{j}\right)  \in\mathcal{B}\right)  $, implying that if the SPs are
similar (r.h.s. in Eq. \ref{equ:refined}), both are of the same label, while
the particular labeling can not be deduced. But, the unary term relates to the
labeling of each particular SP $s_{i}$. Thus, by assuming that the marginal
unary assignment probabilities are independent from the pairwise term%
\begin{multline}
p\left(  \left(  s_{i},s_{j}\right)  \in\mathcal{F}\right)
=\label{equ:refined1}\\
p\left(  \left(  s_{i},s_{j}\right)  \in\mathcal{F}|s_{i}\in\mathcal{F}%
,s_{j}\in\mathcal{F}\right)  p_{u}\left(  s_{i}\in\mathcal{F}\right)
p_{u}\left(  s_{j}\in\mathcal{F}\right)
\end{multline}
where the unary probabilities $p_{u}\left(  \cdot\right)  $ might be given by
the marginals computed in a coarser resolution scale. Equation
\ref{equ:refined1} has a straightforward interpretation, as if both SPs are
\textit{similar} ($\exp(-\frac{D_{KL}^{s}(G_{i}\Vert G_{j})}{\sigma_{p}}%
)\sim1$), and both SPs are of a \textit{particular label }$\mathcal{L}$
($p_{u}\left(  s_{i}\in\mathcal{L}\right)  \sim1$), then we have that the
pairwise probability $p\left(  \left(  s_{i},s_{j}\right)  \in\mathcal{L}%
\right)  \sim1$.

\section{Experimental Results}

\label{sec:experiemental}

The proposed scheme was experimentally verified by applying it to the
state-of-the-art GrabCut, Pascal VOC09 \cite{pascal-voc-2009}\textbf{,} VOC10
and VOC11 datasets, all of which have ground-truth annotations. The GrabCut
dataset uses a semi-interactive setup where the user supplies initial image
markings, while the Pascal datasets are fully unsupervised.

We applied the Watershed algorithm \cite{WatershedB} to the initial
partitioning of the image to SPs. As it utilizes random seeding for
initiation, the resulting output is not repeatable. Hence, we rerun the
segmentation scheme ten times per image and classified pixels as belonging to
a particular class, using majority voting.

Following Section \ref{subsec:implementation}, we computed the pairwise
assignment probabilities, such that the pairwise probabilities is nonzero for
its $m$ most similar neighboring SPs. Thus, insuring that only similar SPs
will affect the segmentation. We used $m=4$ and $m=8$ for the GrabCut, and
PASCAL (VOC09, VOC10, VOC11) simulations, respectively.

\subsection{GrabCut Image Database}

\label{subsec:grabcut results}

The GrabCut database consists of $50$ color images with corresponding
groundtruth annotations, and several user marking conventions. We applied our
approach using the user marking used by Lempitsky et al. \cite{BoxPrior},
defining the foreground's bounding box. The classification of the pixels in
the bounding box is unknown, while those outside of it are known to be
background. The pixels within a ten pixel distance from the bounding box are
used to train the background model.

We compared the GrabCut, GrabCut-Pinpoint \cite{RotherKB, BoxPrior}, and the
TopoCuts \cite{Topological} segmentation schemes, where the results of
\cite{RotherKB,BoxPrior} are cited from \cite{BoxPrior}. The proposed scheme
was applied using the initialization scheme and bounding boxes used in
\cite{BoxPrior}. The results for TopoCuts are cited from \cite{Topological},
where Lasso trimaps were used instead of bounding boxes.

Table \ref{table:GrabCut results} reports the mean relative segmentation error
within the bounding box of our framework. The most accurate segmentation is
achieved by the GrabCut-Pinpoint approach, and we attribute that to the
incorporation of the tightness prior of the bounding box, that provides an
additional segmentation cue. Yet, it might complicate user interaction, and is
inapplicable to automatic segmentation, as in the Pascal datasets (Section
\ref{subsec:pascal results}), where a detector is applied to recover a coarse
estimate of the foreground. The proposed scheme is on par with the GrabCut
approach, as due to the small size of this dataset (50 images over all), and
the high segmentation accuracy rate, the differences between the leading
schemes are essentially negligible. We also cite the results of running the
Matlab-based implementation of GrabCut by Ming Xiumingzhang \cite{GrabcutCode}
that we used in the following simulations to quantify the GrabCut's
sensitivity to parameters, and compare it to the proposed scheme. We note that
Xiumingzhang's implementation, as well as the proposed scheme, do not utilize
the segmentation refinement phase detailed in \cite{RotherKB} and is thus less
accurate by 2\%-3\%. We also report the results of replacing the
PBM\ inference scheme \cite{KellerProbSpec} by Loopy Belief Propagation (LBP),
and Max-Flow \cite{GraphCut} as solvers. \begin{table}[tbh]
\centering%
\begin{tabular}
[c]{|l|l|}\hline
\textbf{Algorithm} & \textbf{Mean Error}\\\hline\hline
GrabCut \cite{RotherKB} & 5.1-5.9\%\\\hline
GrabCut-Reference \cite{GrabcutCode} & 9.1\%\\\hline
GrabCut-Pinpoint \cite{BoxPrior} & \textbf{3.7-4.5\%}\\\hline
TopoCuts\cite{Topological} & 6.3-7.7\%\\\hline
LBP + proposed & 7.3\%\\\hline
Max-Flow + proposed & 10\%\\\hline
Proposed scheme & 5.7\%\\\hline
\end{tabular}
\caption{Segmentation results for the GrabCut dataset. We report the mean
error within the bounding box. We report the results of applying a reference
implementation of GrabCut \cite{GrabcutCode}, and the optimization of the
proposed assignment probabilities using Loopy Belief Propagation (LBP), and
Max-Flow \cite{GraphCut} as solvers.}%
\label{table:GrabCut results}%
\end{table}

\begin{figure}[tbh]
\centering%
\begin{tabular}
[c]{cccc}%
Input image & $K=3$ & $K=4$ & $K=5$\\
\includegraphics[width=0.20\linewidth]{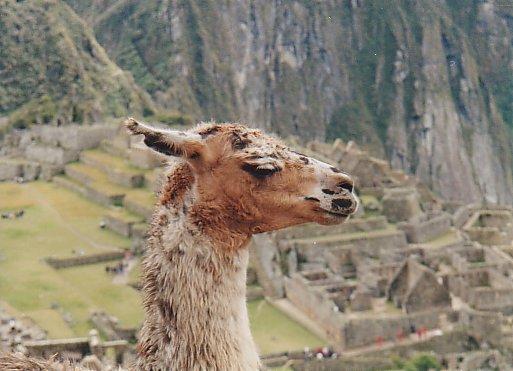} &
\includegraphics[width=0.20\linewidth]{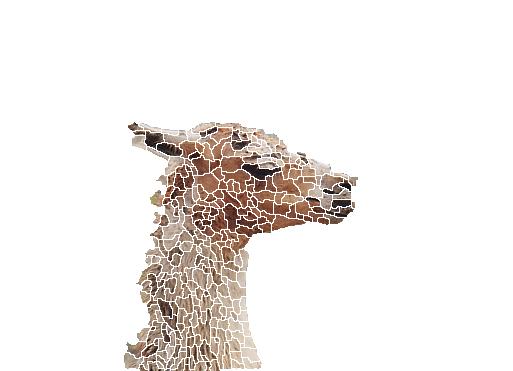} &
\includegraphics[width=0.20\linewidth]{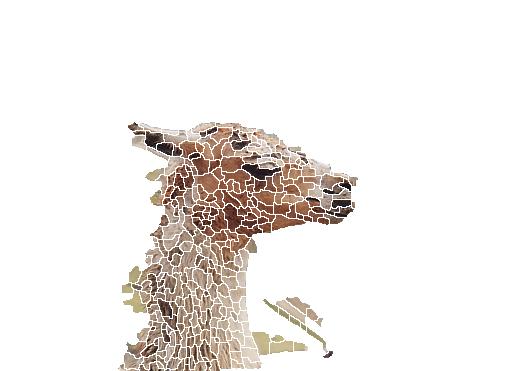} &
\includegraphics[width=0.20\linewidth]{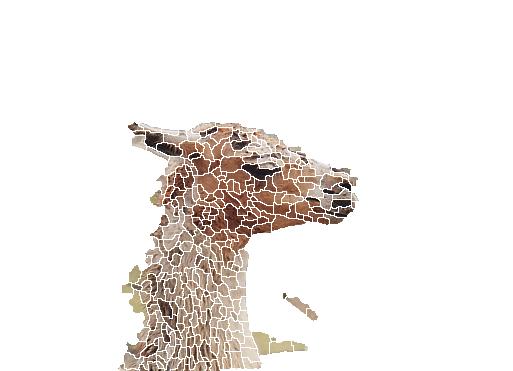}\\
\newline\includegraphics[width=0.20\linewidth]{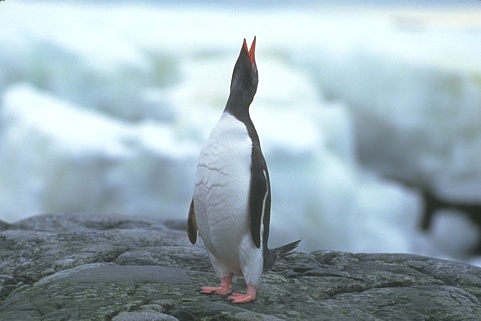} &
\includegraphics[width=0.20\linewidth]{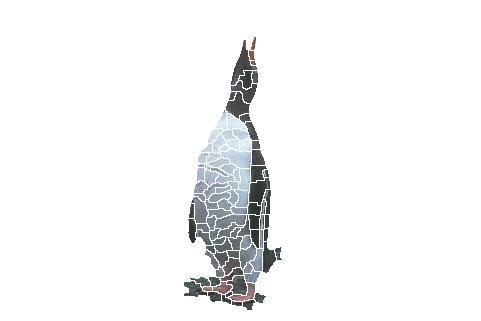} &
\includegraphics[width=0.20\linewidth]{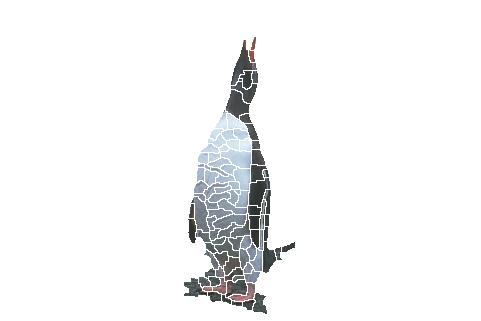} &
\includegraphics[width=0.20\linewidth]{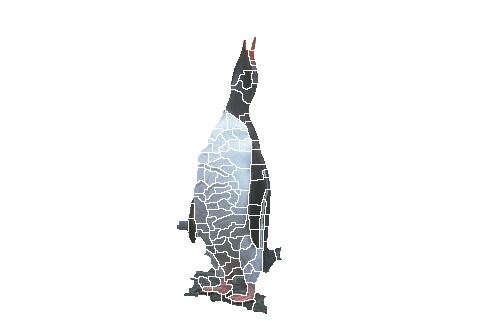}\\
\newline\includegraphics[width=0.20\linewidth]{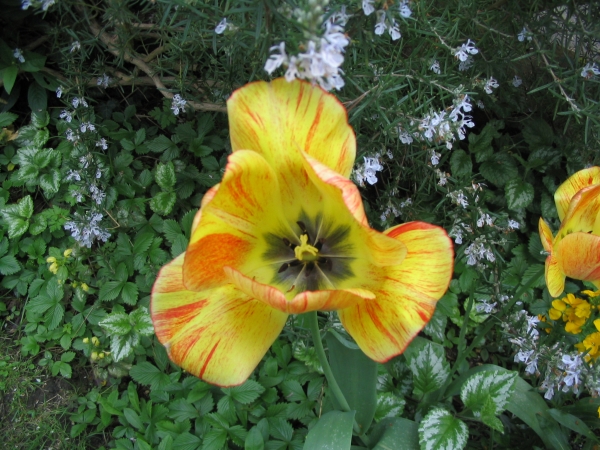} &
\includegraphics[width=0.20\linewidth]{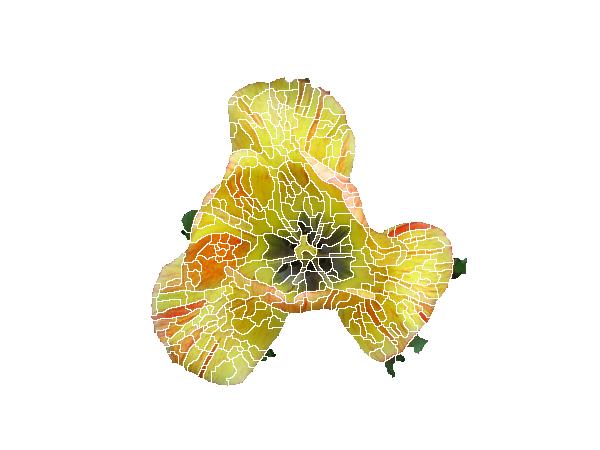} &
\includegraphics[width=0.20\linewidth]{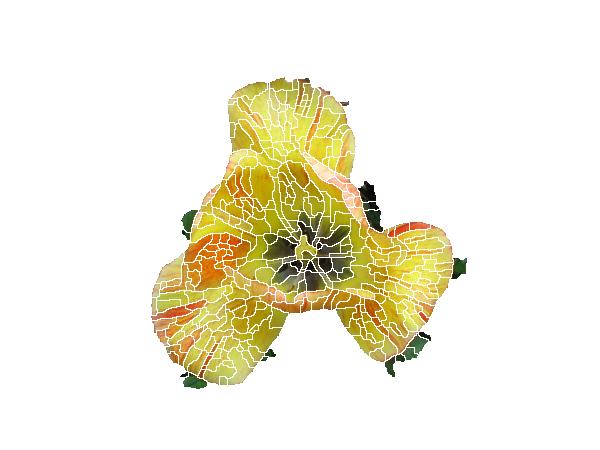} &
\includegraphics[width=0.20\linewidth]{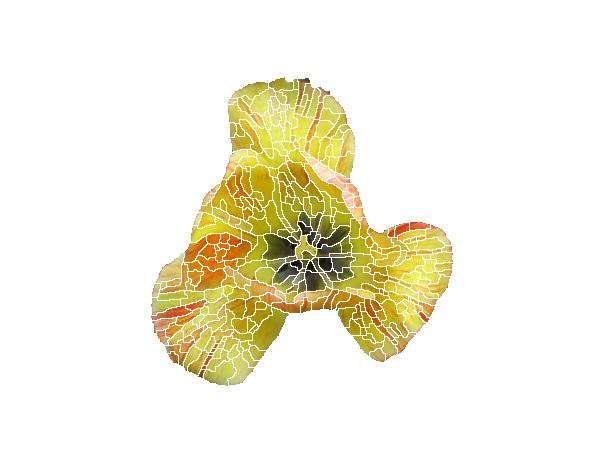}
\end{tabular}
\caption{Segmentation results of the GrabCut dataset for a varying number of
GMM components $K$. The left column depicts the input image.}%
\label{fig:gmm}%
\end{figure}

\begin{figure}[tbh]
\centering
\subfigure[Proposed scheme]{\includegraphics[width=0.80\linewidth]{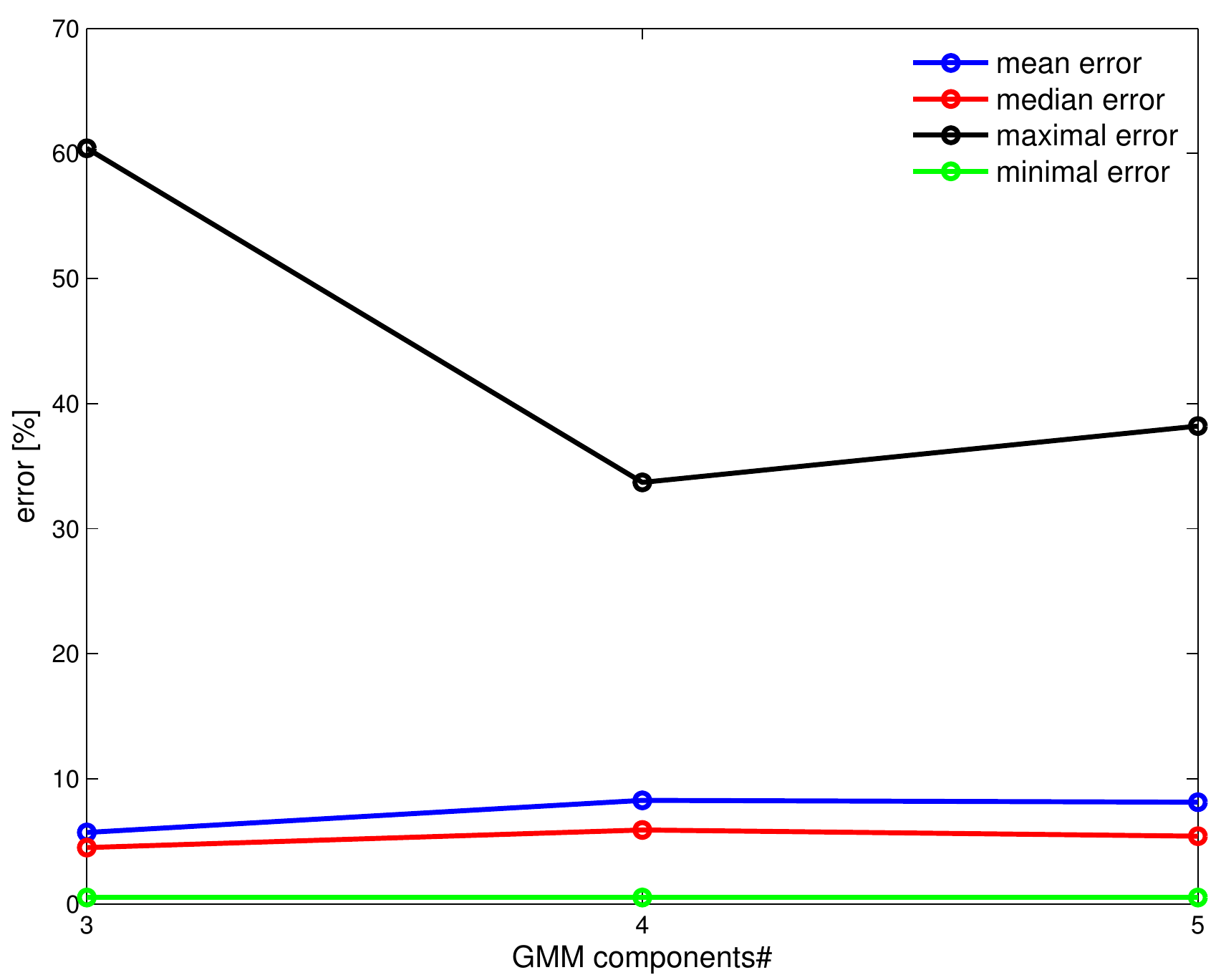}}
\subfigure[GrabCut]{\includegraphics[width=0.80\linewidth]{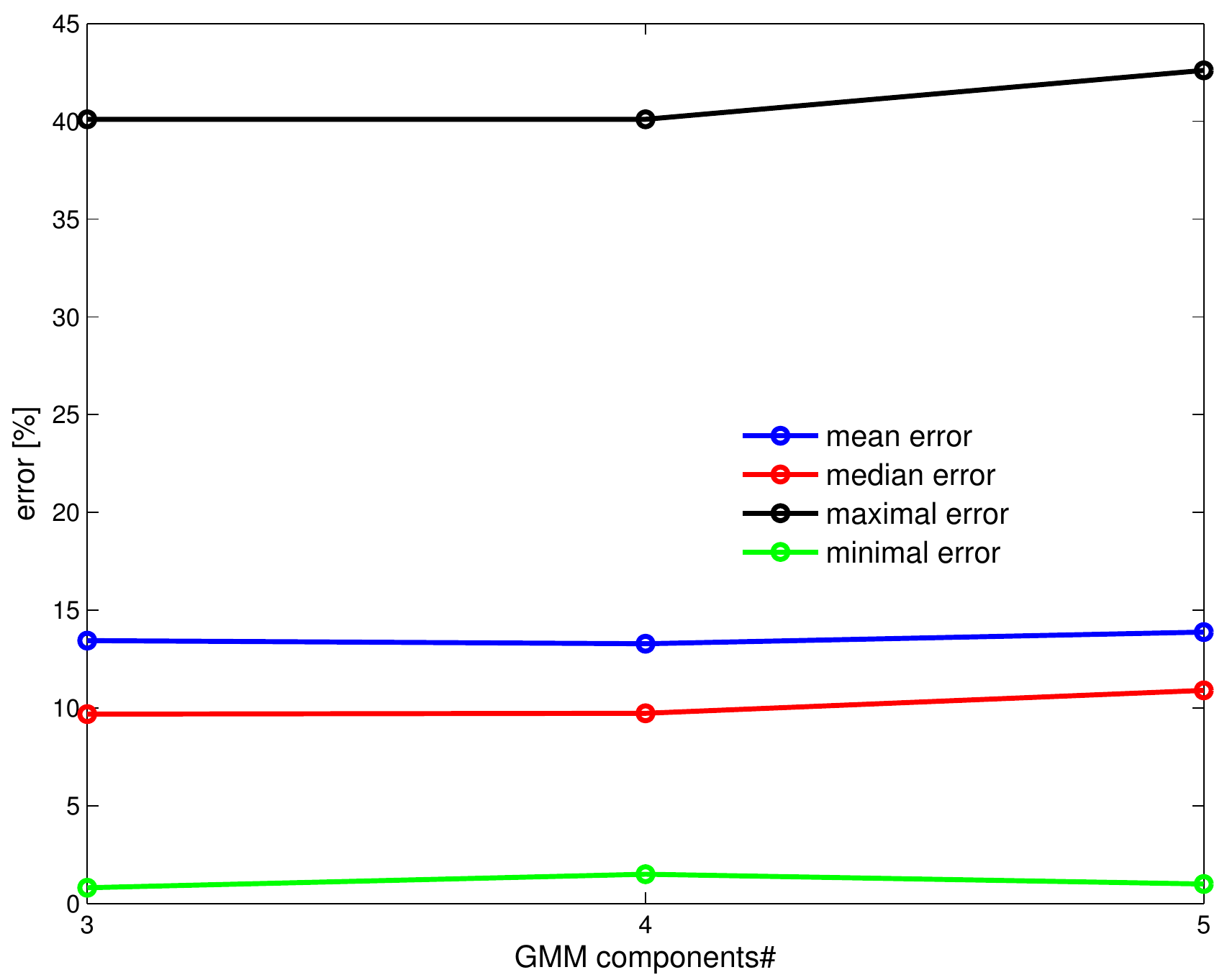}}
\caption{Segmentation results of the GrabCut dataset for a varying number of
GMM components $K=\left\{  3,4,5\right\}  $, for the proposed scheme (a) and
GrabCut scheme (b).}%
\label{fig:accuracy k gmm}%
\end{figure}

The foreground and background are modeled by GMMs with $K=3$ components, which
were shown to yield the highest average accuracy over the entire image set.
Yet, some images are better segmented using a different number of components.
This is depicted in Fig. \ref{fig:gmm} that presents the segmentation results
for $K=\{3,4,5\}$, and it follows that having a data-driven approach to
setting the number of GMM mixtures, might significantly improve the
segmentation results for certain images. The \textit{average} segmentation accuracy with respect to $K$ is
depicted in Fig. \ref{fig:accuracy k gmm}\textbf{, }for the proposed scheme
and the GrabCut approach. It follows that the average accuracy is insensitive
to the choice of $K$, and the mean accuracy of the proposed scheme is shown to
outperform the GrabCut segmentation accuracy.

\begin{figure}[tbh]
\centering\subfigure[Proposed scheme]{\includegraphics[width=0.80\linewidth]{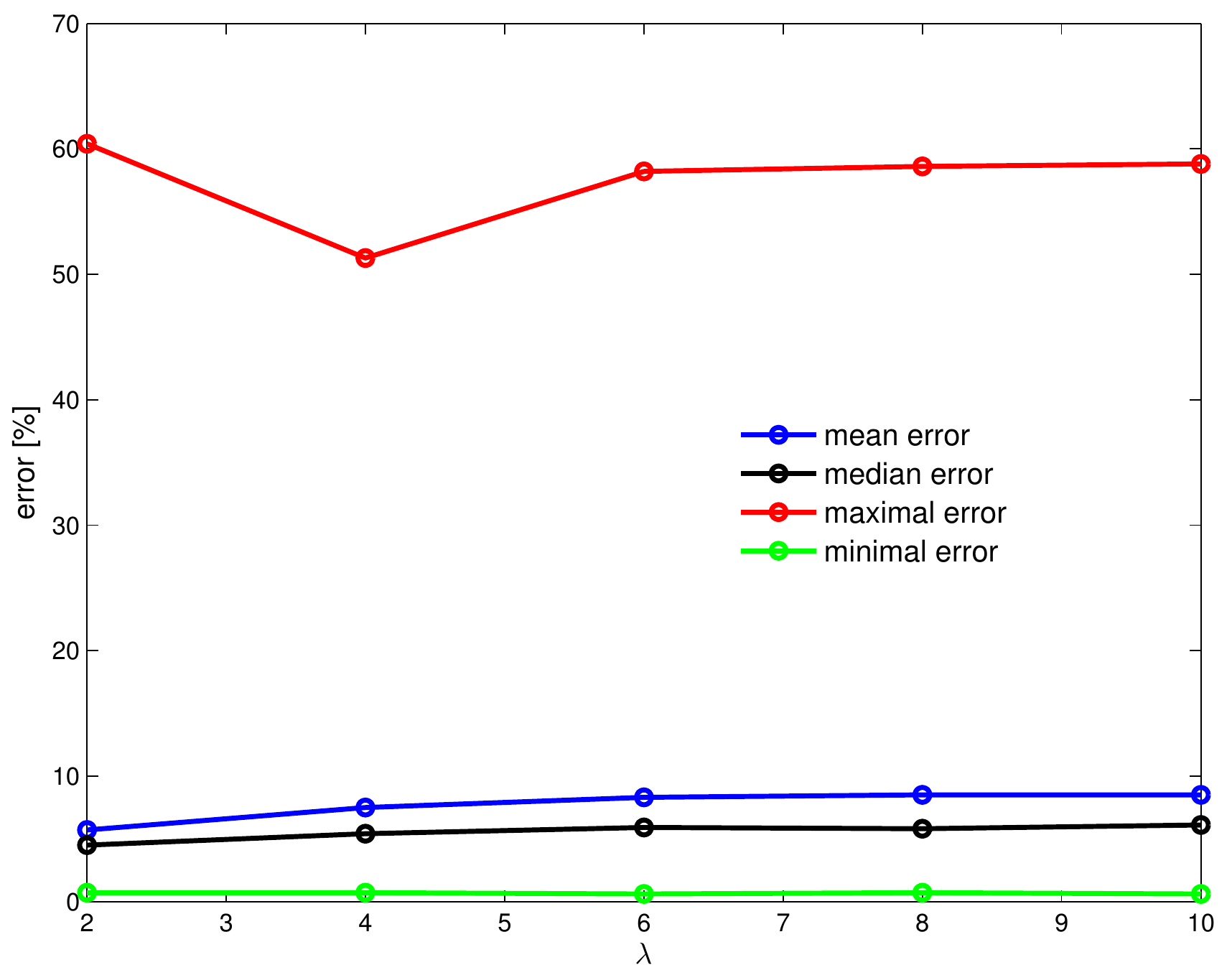}}
\subfigure[GrabCut]{\includegraphics[width=0.80\linewidth]{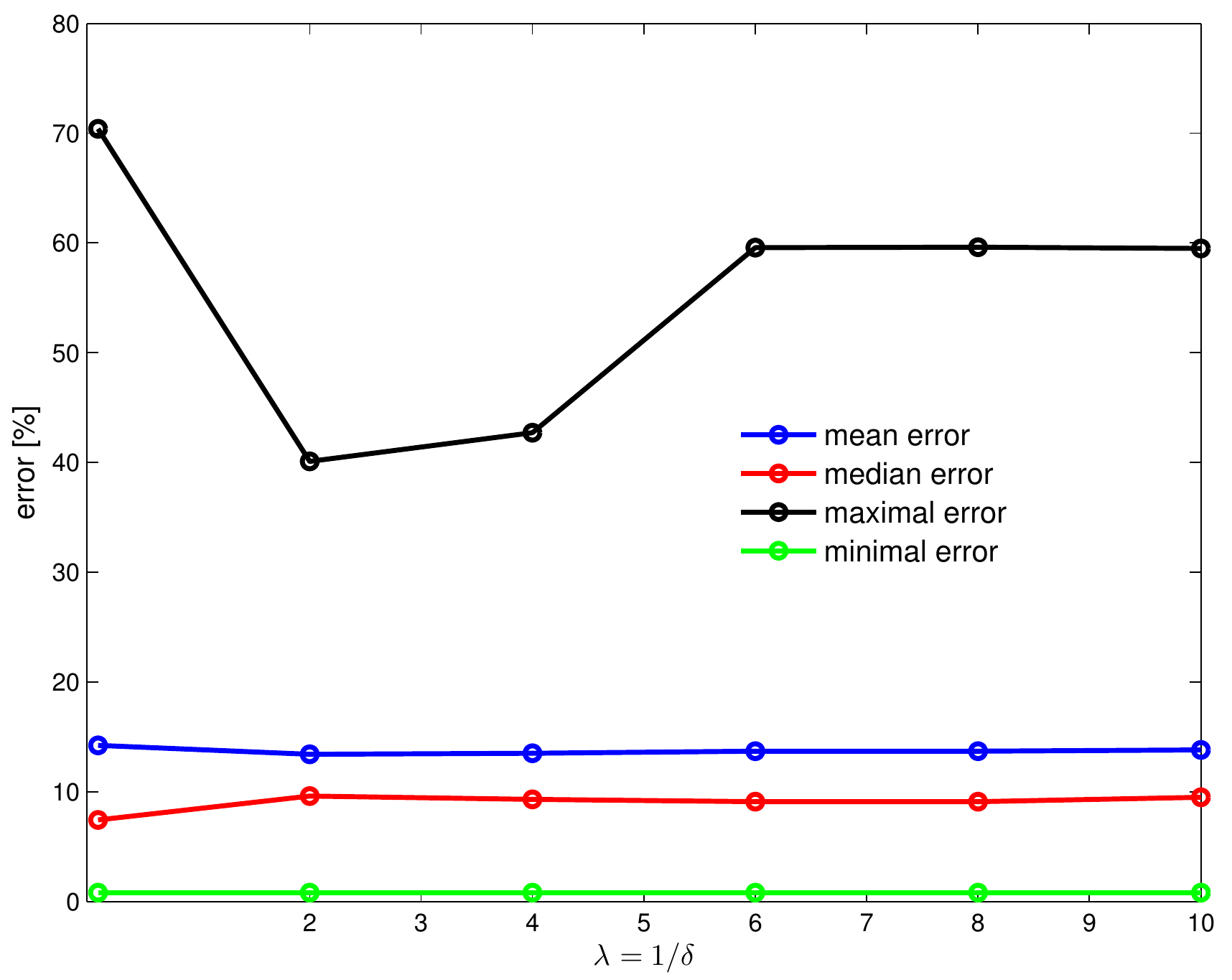}}\caption{Segmentation
accuracy of the GrabCut dataset using varying $\lambda$, the relative
weighting of the unary and binary terms, for the proposed scheme (a) and
GrabCut scheme (b).}%
\label{fig:accuracy lamda}%
\end{figure}

We also studied the accuracy sensitivity with respect to $\lambda$, the
relative weighting of the unary and binary terms, in Fig.
\ref{fig:accuracy lamda}. The average accuracy seems insensitive to the choice
of $\lambda$, similar to the results in Fig. \ref{fig:accuracy k gmm}, where
the proposed scheme outperforms the GrabCut. We note that due to the different
potentials and weights used by the proposed scheme and GrabCut, respectively,
we did not use the same values of $\lambda$ for both schemes. We varied
$\lambda$ for both schemes around their default values.

\begin{figure}[tbh]
\centering%
\begin{tabular}
[c]{cccc}%
$\lambda=2$ & $\lambda=4$ & $\lambda=6$ & $\lambda=10$\\
\includegraphics[width=0.20\linewidth]{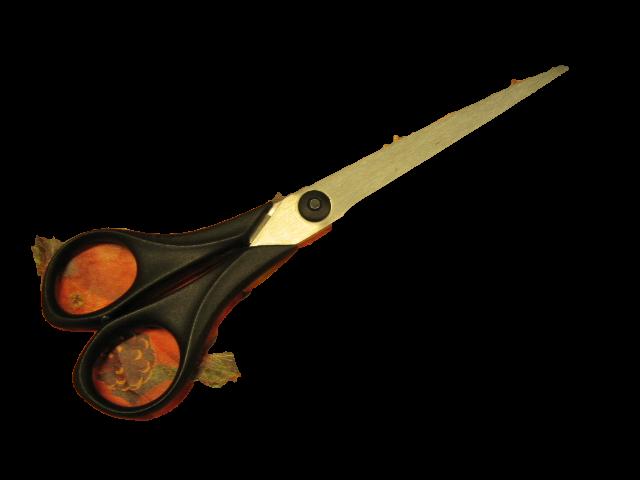} &
\includegraphics[width=0.20\linewidth]{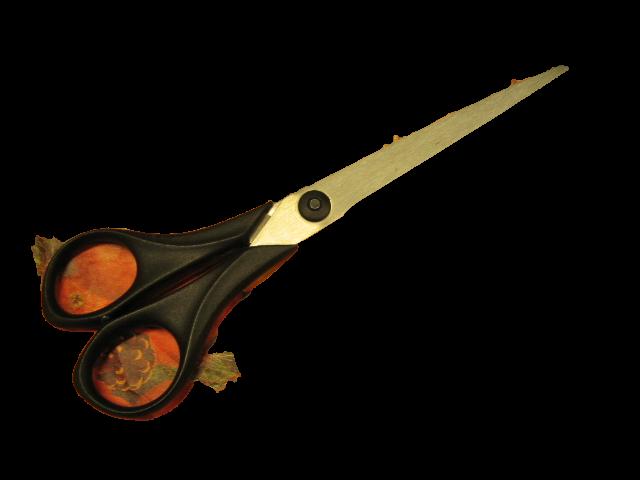} &
\includegraphics[width=0.20\linewidth]{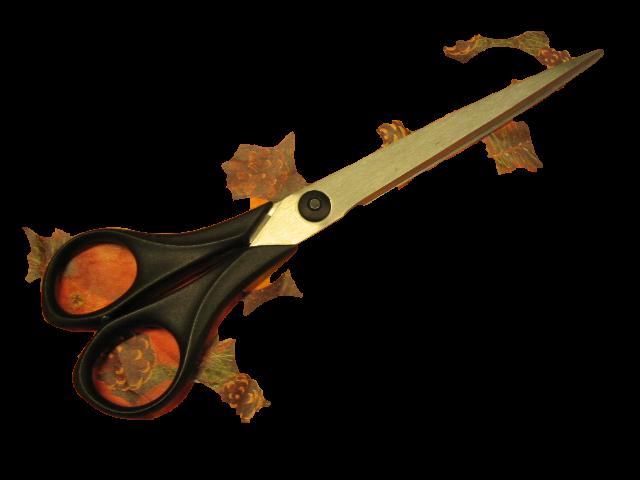} &
\includegraphics[width=0.20\linewidth]{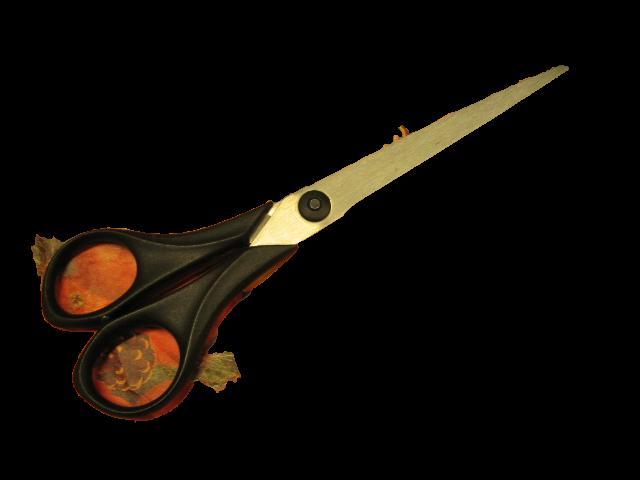}\\
\newline\includegraphics[width=0.20\linewidth]{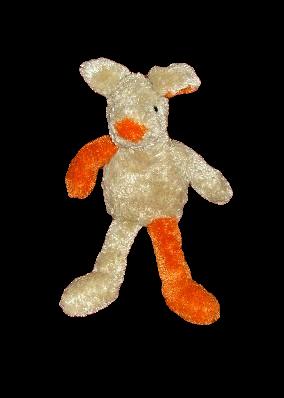} &
\includegraphics[width=0.20\linewidth]{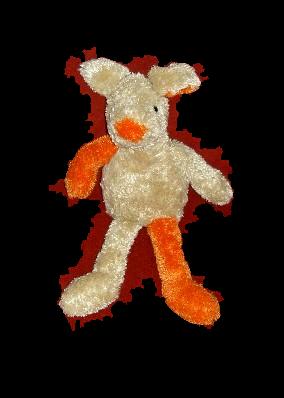} &
\includegraphics[width=0.20\linewidth]{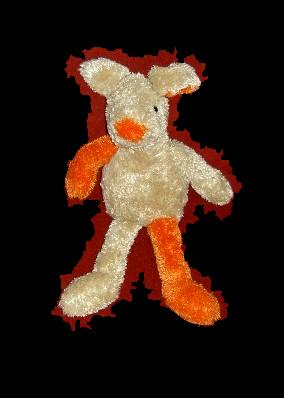} &
\includegraphics[width=0.20\linewidth]{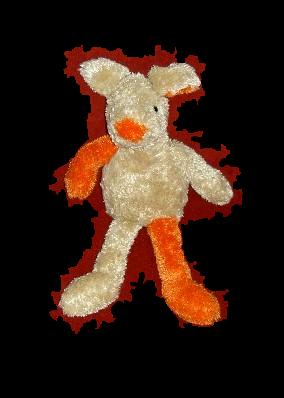}\\
\newline\includegraphics[width=0.20\linewidth]{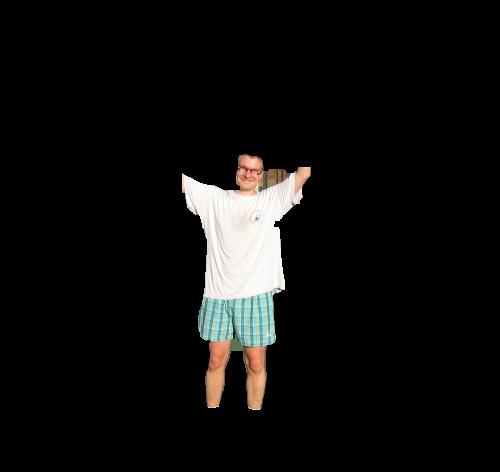} &
\includegraphics[width=0.20\linewidth]{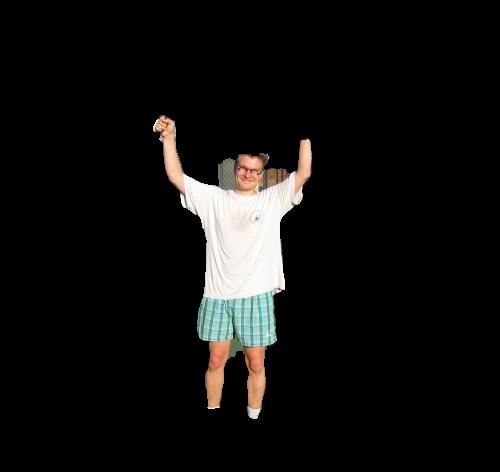} &
\includegraphics[width=0.20\linewidth]{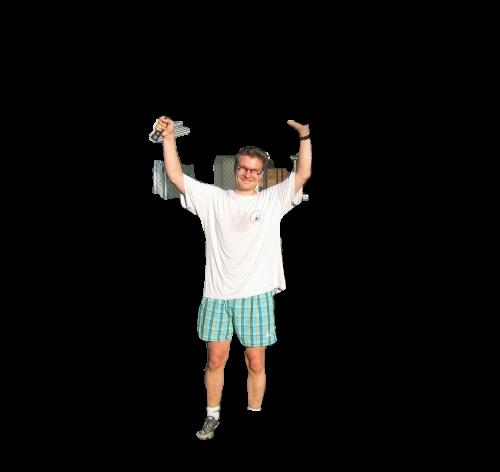} &
\includegraphics[width=0.20\linewidth]{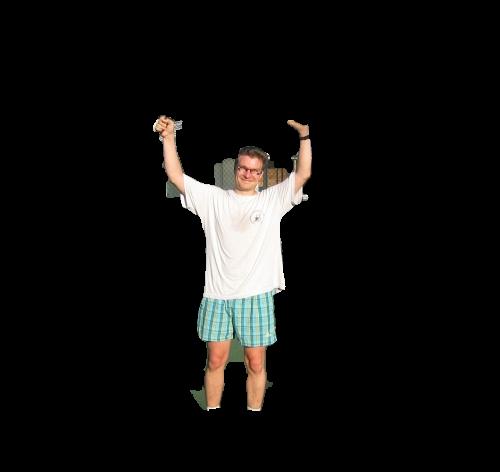}
\end{tabular}
\caption{Segmentation results of the GrabCut dataset for varying weighting
tradeoffs $\lambda$ between the unary and binary terms. The larger $\lambda$
the more emphasis is given to the unary term in Eq. \ref{equ:lamda}.}%
\label{fig:image lamda}%
\end{figure}

Following Eq. \ref{equ:lamda}, the larger $\lambda,$ the more relative weight
is given to the unary probabilities that encode the $\mathcal{F}$%
/$\mathcal{B}$ color models. In contrast, the pairwise assignment
probabilities encode the local classification structure. This is exemplified
in Fig. \ref{fig:image lamda} where we depict the visual outcome of varying
$\lambda$. For $\lambda=2$ the pairwise probabilities dominate the
segmentation, resulting in accurate boundaries, while as $\lambda$ increases
the segmentation becomes less accurate.

\begin{figure}[tbh]
\centering\subfigure[Varying $\lambda$]{\includegraphics[width=0.80\linewidth]{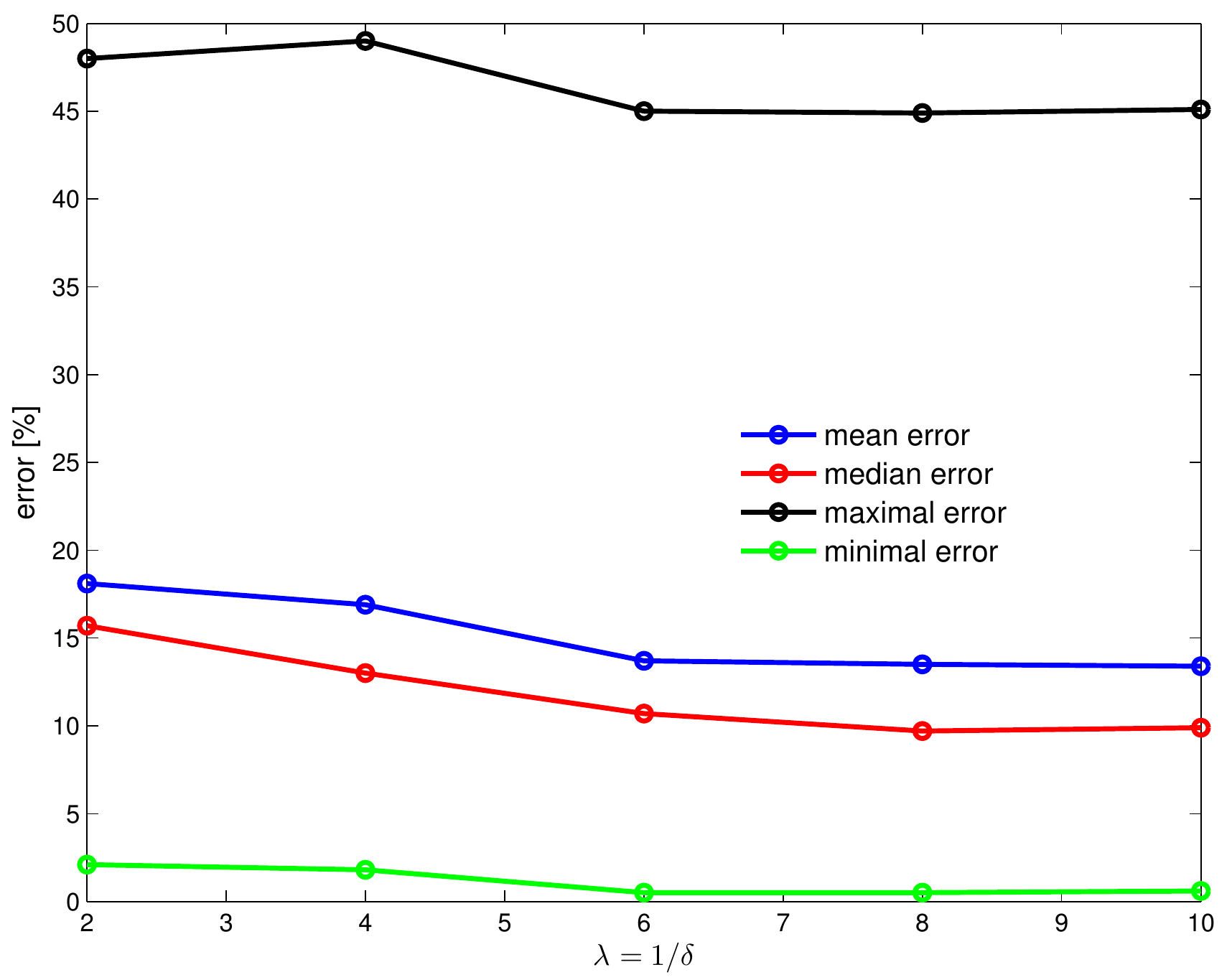}}
\subfigure[Varying $K$]{\includegraphics[width=0.80\linewidth]{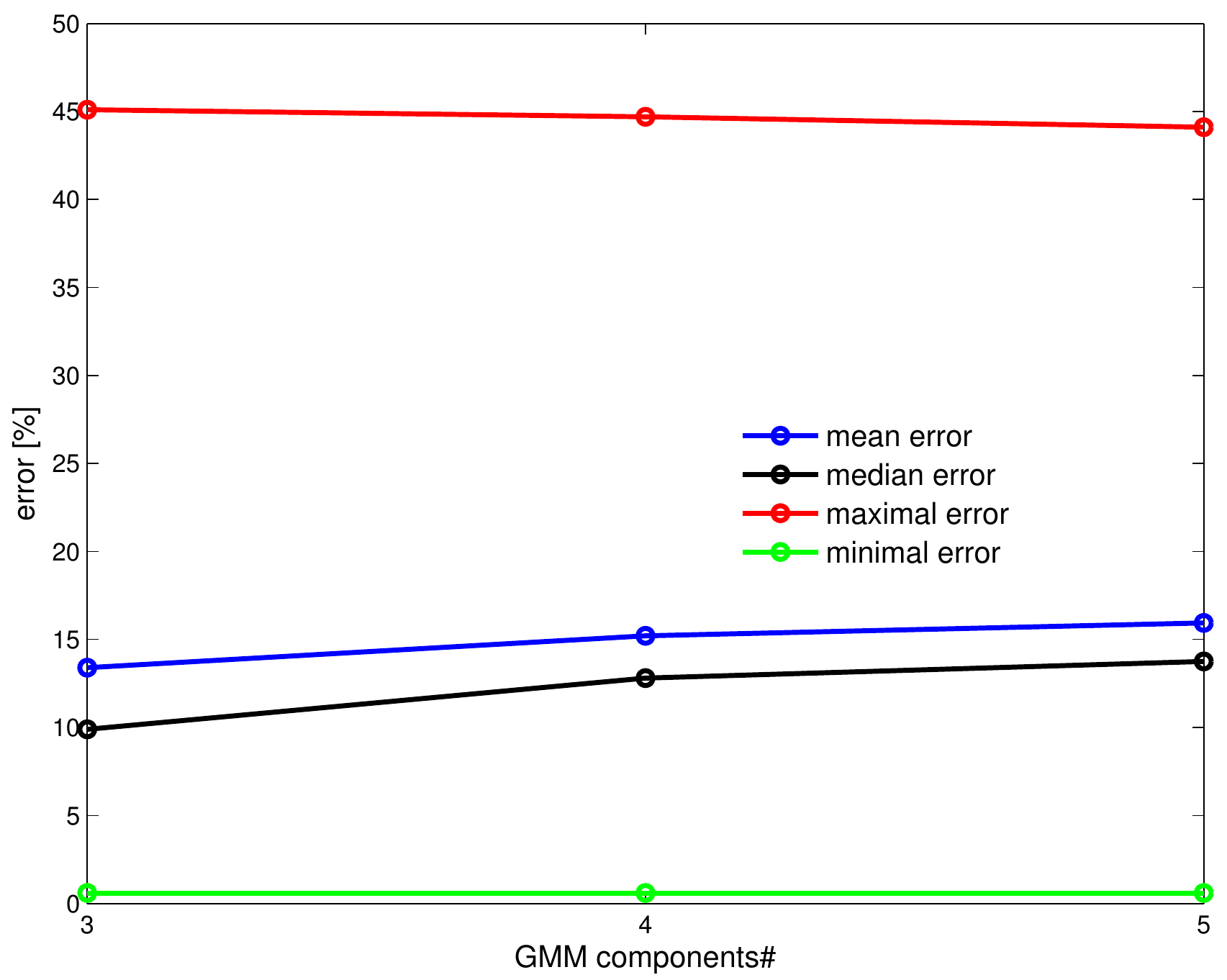}}\caption{Image
segmentation using the max-flow solver and the proposed probabilities as
potentials. (a) Segmentation accuracy with respect to the relative weighting
$\lambda$ using GMMs with $K=3$, and with respect to a varying number of GMM
components in (b) .}%
\label{fig:GrabCut solver results}%
\end{figure}

In order to quantify the relative influence of the proposed inference scheme
(Section \ref{subsec:assignment}) and the proposed unary and pairwise terms
(Sections \ref{subsec:assignment} and \ref{subsec:gmm}, respectively), on the
segmentation accuracy, we reformulated the unary and pairwise terms as
weights, and applied a state-of-the-art max-flow/min-cut solver
\footnote{http://vision.csd.uwo.ca/wiki/vision/upload/d/d7/Bk\_matlab.zip}.
The unary and binary weights are given by%
\begin{equation}
\phi\left(  s_{i}\right)  =1-p_{u}\left(  s_{i}\in\mathcal{F}\right)  ,
\end{equation}
and%
\begin{equation}
\phi\left(  s_{i},s_{j}\right)  ={2 \cdot} p\left(  s_{i}\in\mathcal{F},s_{j}%
\in\mathcal{B}\right)  ,
\end{equation}
respectively, where $p_{u}\left(  \cdot\right)  $ and $p\left(  \cdot
,\cdot\right)  $ are the unary and pairwise probabilities defined in Sections
\ref{subsec:assignment} and \ref{subsec:gmm}. Figures
\ref{fig:GrabCut solver results}a and \ref{fig:GrabCut solver results}b depict
the results for a varying number of GMM components $K$ and relative weighting
$\lambda$, respectively. It follows that using the proposed probabilities with
the max-flow/min-cut solver results in lower segmentation
accuracy.

\begin{figure}[tbh]
\centering\subfigure[Varying $\lambda$]{\includegraphics[width=0.80\linewidth]{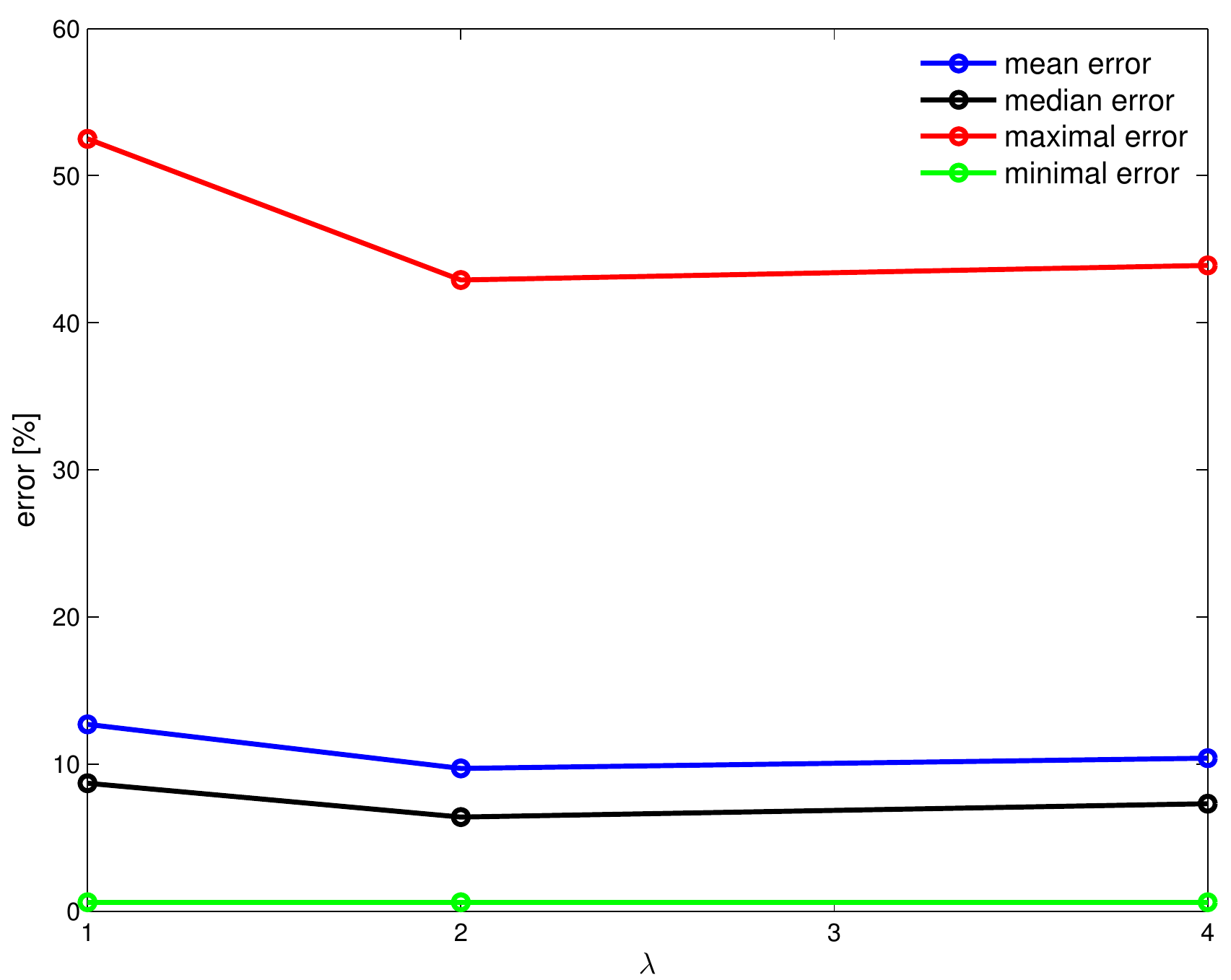}}
\subfigure[Varying $K$]{\includegraphics[width=0.80\linewidth]{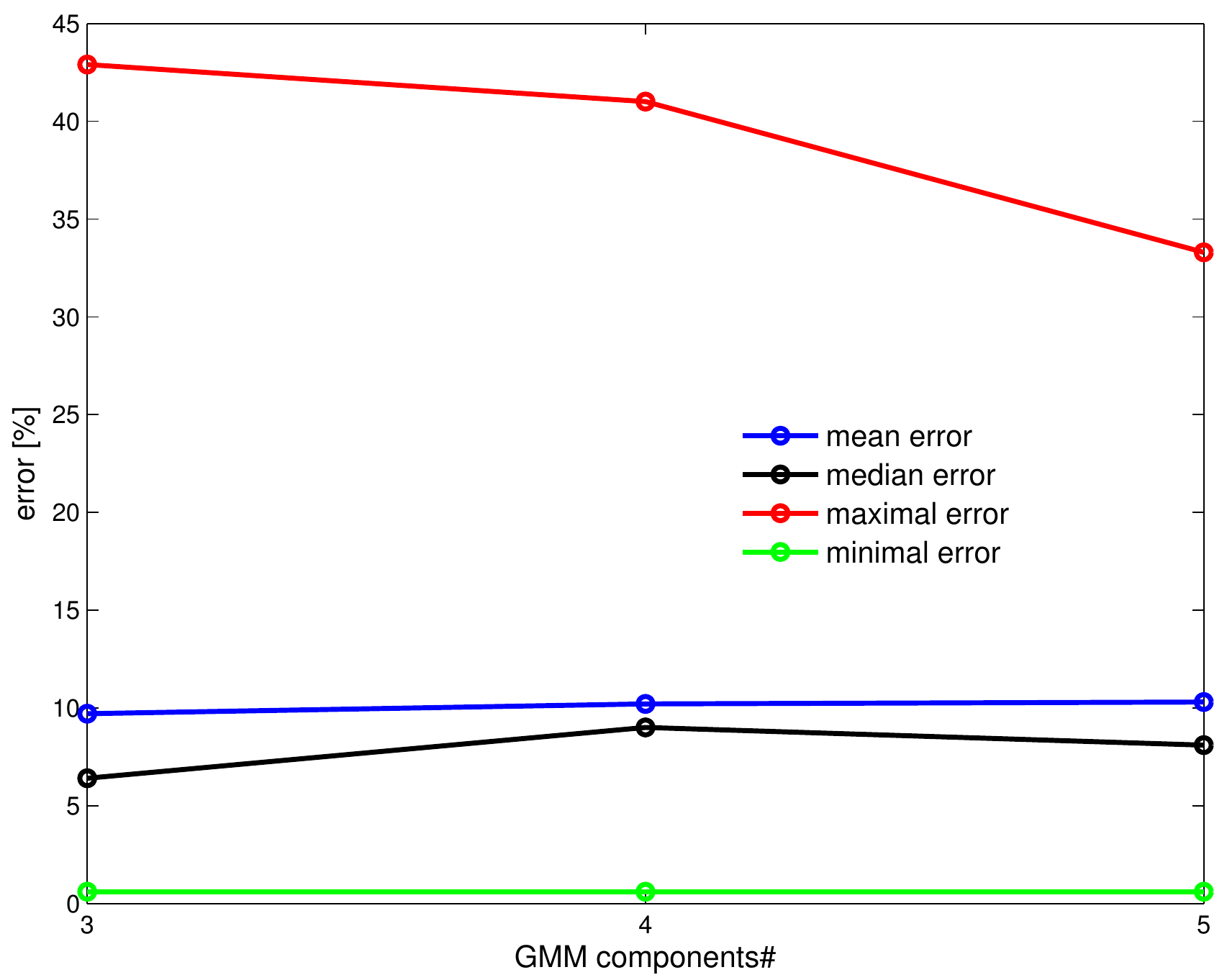}}\caption{Image
segmentation using the Loopy Belief Propagation solver and the proposed
probabilities as potentials. (a) Segmentation accuracy with respect to the
relative weighting $\lambda$ using GMMs with $K=3$. \ (b) Segmentation
accuracy with respect to the number of GMM\ components $K$ and $\lambda=2.$}%
\label{fig:LBP solver results}%
\end{figure}

We also applied the Loopy Belief Propagation (LBP) as a solver utilizing the
unary and binary probabilities as in Fig. \ref{fig:GrabCut solver results}.
For that we utilized the UGM
toolbox\footnote{https://www.cs.ubc.ca/\symbol{126}schmidtm/Software/UGM.html}%
, and the results are shown in Fig. \ref{fig:LBP solver results}. The
LBP-based solver relates to the proposed scheme as both are probabilistic
inference schemes, in contrast to the MaxFlow \cite{GraphCut} that is a
discrete optimization scheme. The main difference being that the PGM
\cite{KellerProbSpec} is able to handle probabilities represented by dense
graphs, while the LBP is optimal for trees and might diverge when applied to
dense probabilistic graphs. The inference graph in image segmentation is
typically sparse, as the pairwise probabilities only relate to neighboring
SPs, thus allowing the LBP to converge. In our simulations depicted in Fig.
\ref{fig:LBP solver results}, the LBP-based solver yielded an accuracy 7.3\%,
that is inferior to the proposed scheme.

\begin{figure}[tbh]
\centering%
\begin{tabular}
[c]{cccc}%
\includegraphics[width=0.15\linewidth]{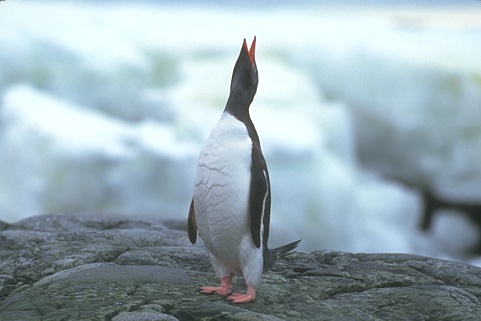} &
\includegraphics[width=0.20\linewidth]{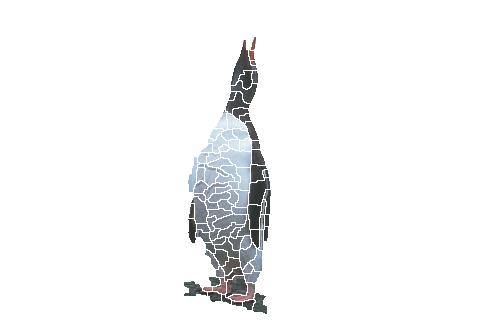} &
\includegraphics[width=0.20\linewidth]{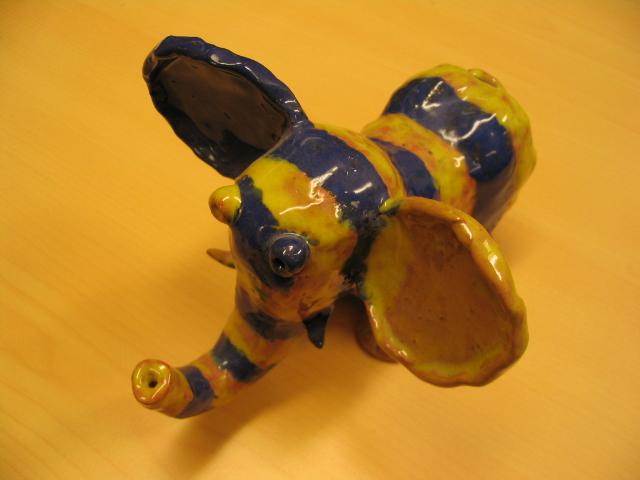} &
\includegraphics[width=0.20\linewidth]{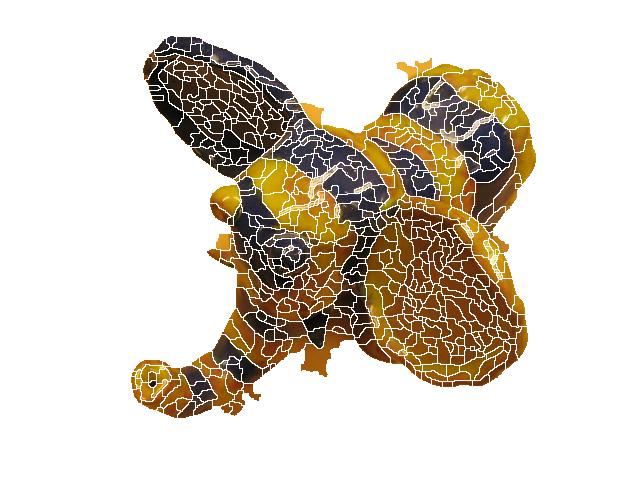}\\
\newline\includegraphics[width=0.20\linewidth]{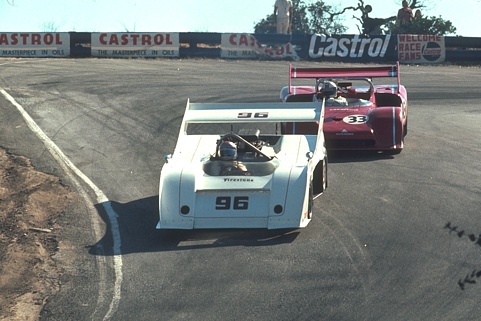} &
\includegraphics[width=0.20\linewidth]{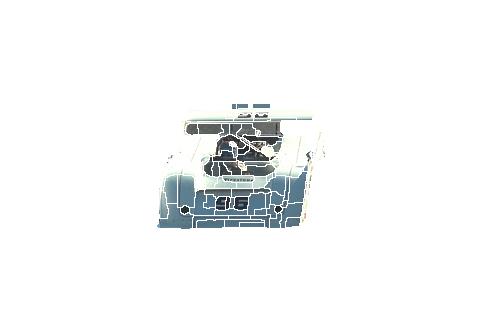} &
\includegraphics[width=0.20\linewidth]{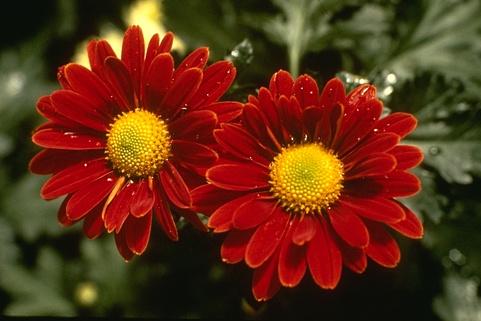} &
\includegraphics[width=0.20\linewidth]{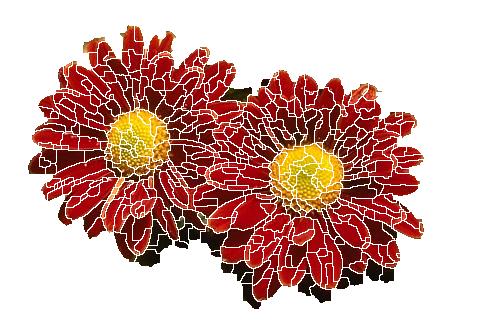}\\
\newline\includegraphics[width=0.20\linewidth]{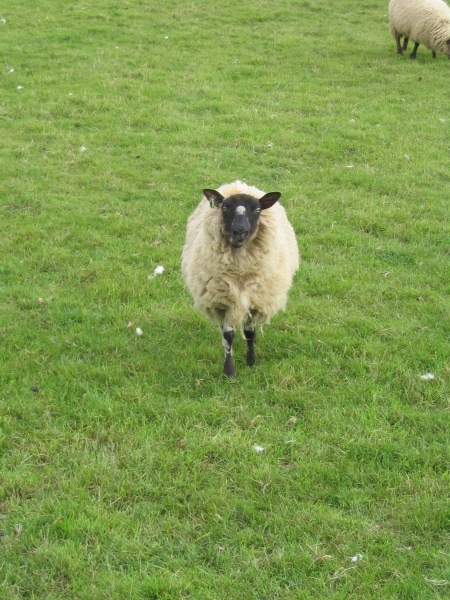} &
\includegraphics[width=0.20\linewidth]{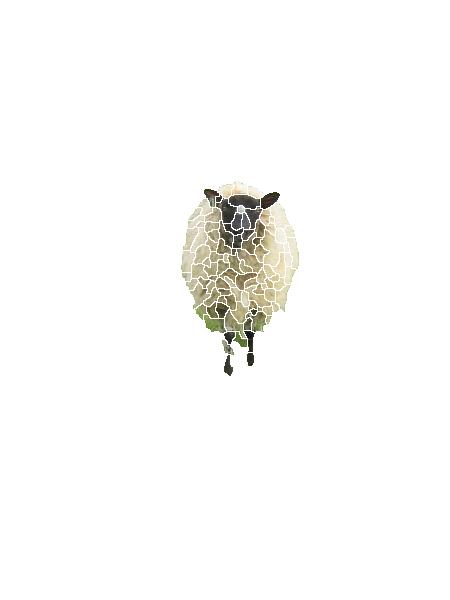} &
\includegraphics[width=0.20\linewidth]{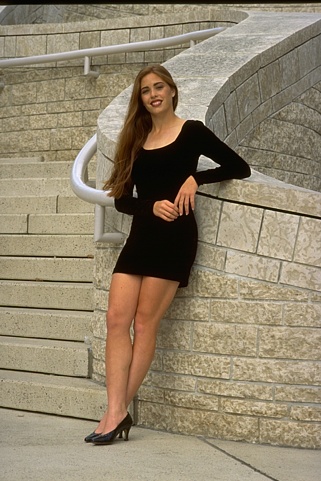} &
\includegraphics[width=0.20\linewidth]{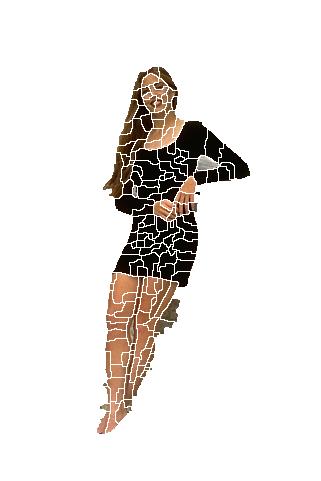}\\
\newline\includegraphics[width=0.20\linewidth]{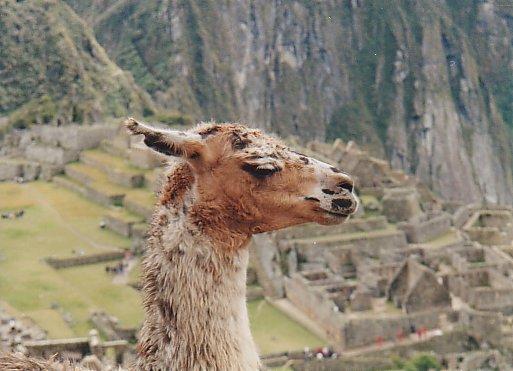} &
\includegraphics[width=0.20\linewidth]{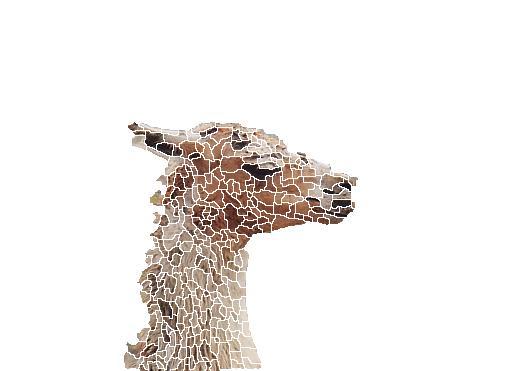} &
\includegraphics[width=0.20\linewidth]{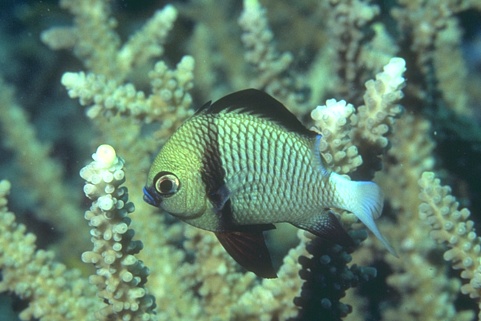} &
\includegraphics[width=0.20\linewidth]{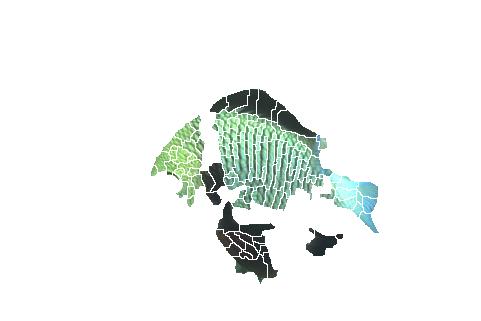}\\
\includegraphics[width=0.20\linewidth]{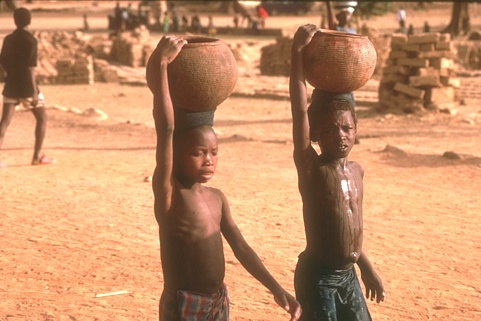} &
\includegraphics[width=0.20\linewidth]{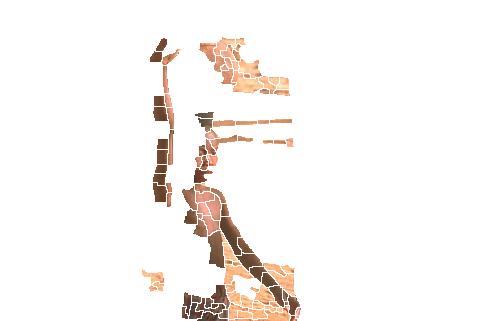} &  &
\end{tabular}
\caption{Segmentation results of the GrabCut dataset, using GMMs with $K=3$
components, and $N=3$ KNN for the pairwise assignment probabilities. }%
\label{fig:results}%
\end{figure}

We present additional qualitative segmentation results of the proposed scheme
in Fig. \ref{fig:results} 

\subsection{Pascal VOC Datasets}

\label{subsec:pascal results}

The Pascal VOC09 \cite{pascal-voc-2009}, Pascal VOC10 \cite{pascal-voc-2010}
and PASCAL VOC11 \cite{pascal-voc-2011} datasets consist of 1499, 1928 and
2223 images, respectively. The PASCAL dataset groundtruth annotation consists
of the annotations of 20 classes of objects, and we considered each annotated
pixel as belonging to the foreground.

In this dataset the TriMap initialization set is not utilized, as in the
GrabCut database, and is segmented in automatically. To add an initialization
step, that provides a coarse estimate of the segmentation, we implemented the
segmentation prior proposed by Rosenfeld et al. \cite{geometric} and detailed
in Section \ref{subsection:Unsupervised}, using $p_{0}=0.4$ as the background
probability detection threshold. For each dataset, we randomly chose half of
the images to train the detector. The background pixels closest to the
unclassified portion of the image were used for background model training,
similar to the user provided input in the semi-{automatic} formulation of our
approach, and the proposed scheme was applied as in Section
\ref{subsec:grabcut results}. We experimentally used $K=5$ GMM components for
background modeling, and $K=3$ for foreground modeling, while the RBF
bandwidths are auto-tuned using Eqs. \ref{equ:sigmaub} and \ref{equ:sigmapb}.

We segment the image into $\mathcal{F}$/$\mathcal{B}$ regions, considering all
the objects in the image as being foreground, and quantify the segmentation
accuracy using the \textit{overlap score} \cite{geometric}
\begin{equation}
overlap=|\frac{S\cap S^{\prime}}{S\cup S^{\prime}}|
\end{equation}
where $S$ is the ground truth and $S^{\prime}$ is the segmentation result.

Table \ref{table:voc results} compares the mean overlap score for the
segmentation of the Pascal VOC09, Pascal VOC10 and Pascal VOC11 datasets using
the proposed scheme, {CPMC \cite{CPMC2} and} the work of Kuettel et
al. \cite{sliding_window}. We also compare against Global Image Neighbor
Transfer (GINT) and GrabCut 50\% (GC50) that were used as a baseline by
Kuettel et al. \cite{sliding_window}. In GINT, the $m=5$  training
images {most similar} to the test image, (according to the GIST descriptor) were averaged and
thresholded to produce the final segmentation. In GC50, a patch consisting of
50\% of the image size, located at the image center, was used as an initial
estimate of the foreground.

We also compared against a variant of the proposed scheme, where the initial
unary probabilities are estimated using only the background. The gist of this
formulation is that the initial \textit{unknown region} is known to consist of
both $\mathcal{F}$ and $\mathcal{B}$ SPs, while the initial background region
only consists of $\mathcal{B}$ SPs. The initial estimate of \ the GMMs is thus
given by
\begin{align}
p_{u}\left(  \mathbf{s}_{i}\in\mathcal{B}\right)   &  =\exp\left(
-\frac{\tilde{D}_{KL}(G_{i}\Vert GMM_{\mathcal{B}})}{\sigma_{u}}\right)
\newline\newline,\\
p_{u}\left(  \mathbf{s}_{i}\in\mathcal{F}\right)   &  =1-p_{u}\left(
\mathbf{s}_{i}\in\mathcal{B}\right)
\end{align}
where $\sigma_{u}$ is estimated by%
\begin{equation}
\sigma_{u}=\underset{G_{i}\in\mathcal{B}}{median}\tilde{D}_{KL}(G_{i}\Vert
GMM_{\mathcal{B}}).
\end{equation}

The remainder of the iterative refinement is applied as in Section
\ref{subsec:refinement}.\begin{table}[tbh]
\centering%
\begin{tabular}
[c]{|c|c|c|c|}\hline
\textbf{Method} & \textbf{VOC 2009} & \textbf{VOC 2010} & \textbf{VOC
2011}\\\hline\hline
\multicolumn{1}{|l|}{GC50\cite{sliding_window}} & - & 0.30 & -\\\hline
\multicolumn{1}{|l|}{GINT\cite{sliding_window}} & - & 0.27 & -\\\hline
\multicolumn{1}{|l|}{CPMC \cite{CPMC2} (best K)} & - & 0.34 & -\\\hline
\multicolumn{1}{|l|}{Kuettel et al \cite{sliding_window}} & - & \textbf{0.48}
& -\\\hline
\multicolumn{1}{|l|}{Proposed scheme} & 0.46 & 0.47 & 0.47\\\hline
\multicolumn{1}{|l|}{Proposed scheme-gb} & \textbf{0.47} & \textbf{0.48} &
\textbf{0.48}\\\hline
\end{tabular}
\caption{Segmentation results for the Pascal VOC09, VOC2010 and VOC11 datasets
using the overlap score. The higher the overlap score, the better the
segmentation accuracy.}%
\label{table:voc results}%
\end{table}

\begin{figure}[tbh]%
\begin{tabular}
[c]{cccc}%
\subfigure[]{\includegraphics[width=0.20\linewidth]{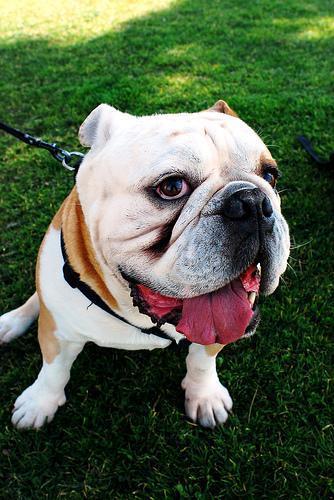}} &
\subfigure[]{\includegraphics[width=0.20\linewidth]{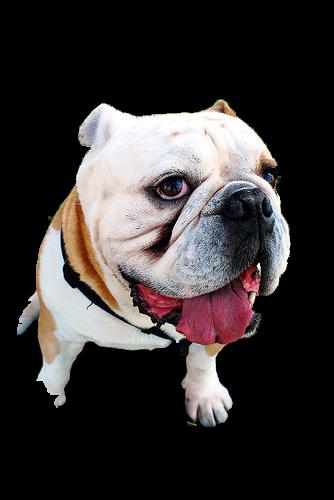}} &
\subfigure[]{\includegraphics[width=0.20\linewidth]{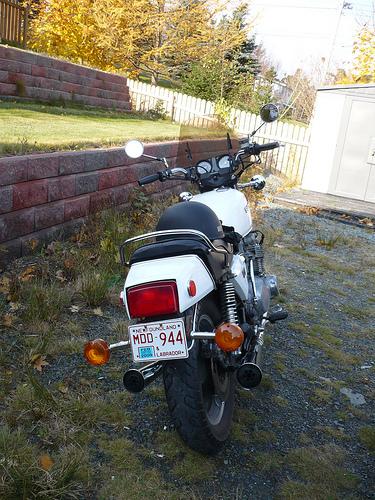}} &
\subfigure[]{\includegraphics[width=0.20\linewidth]{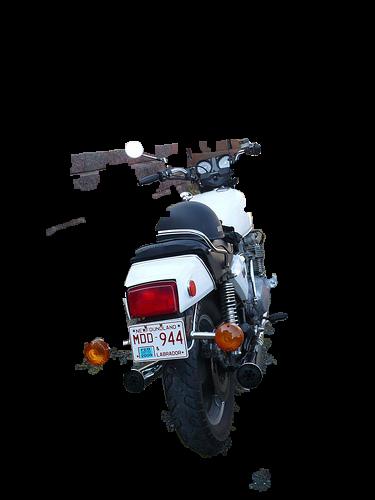}}\\
\newline\subfigure[]{\includegraphics[width=0.20\linewidth]{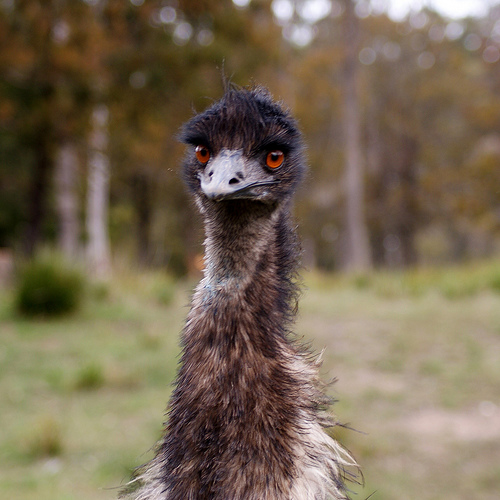}} &
\subfigure[]{\includegraphics[width=0.20\linewidth]{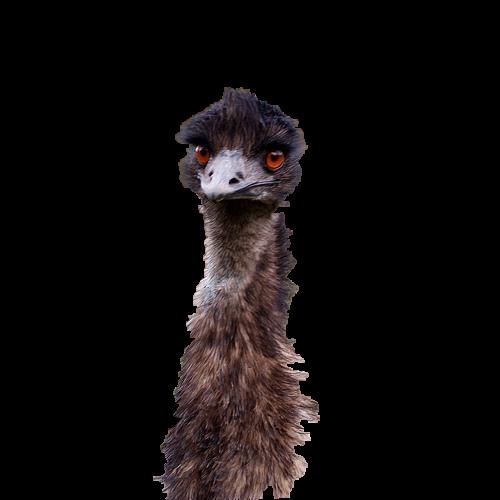}} &
\subfigure[]{\includegraphics[width=0.20\linewidth]{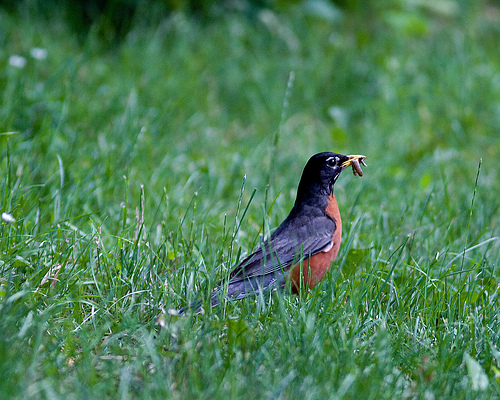}} &
\subfigure[]{\includegraphics[width=0.20\linewidth]{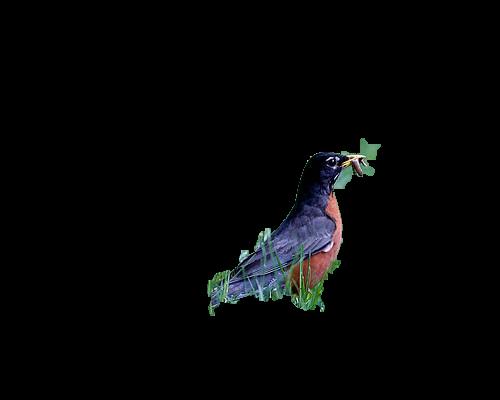}}\\
\newline\subfigure[]{\includegraphics[width=0.20\linewidth]{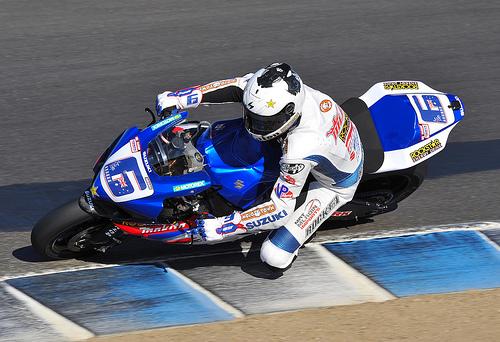}} &
\subfigure[]{\includegraphics[width=0.20\linewidth]{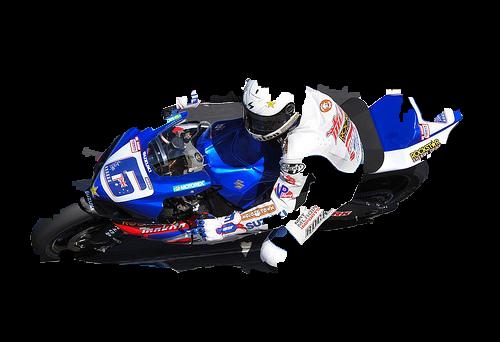}} &
\subfigure[]{\includegraphics[width=0.20\linewidth]{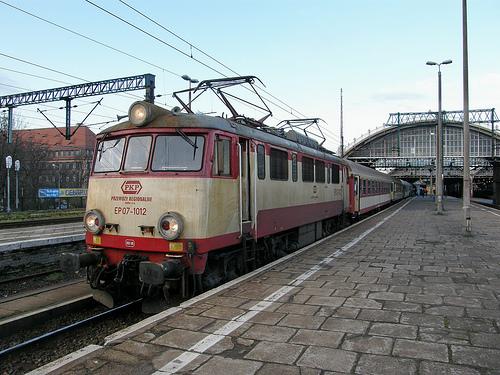}} &
\subfigure[]{\includegraphics[width=0.20\linewidth]{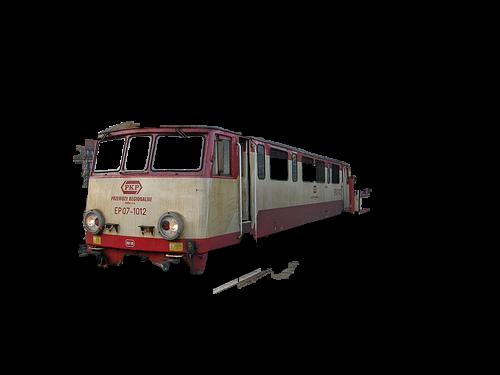}}\\
\newline\subfigure[]{\includegraphics[width=0.20\linewidth]{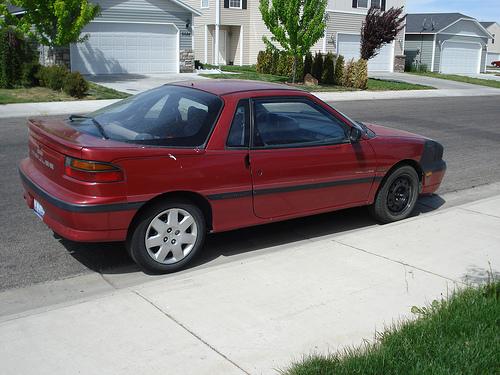}} &
\subfigure[]{\includegraphics[width=0.20\linewidth]{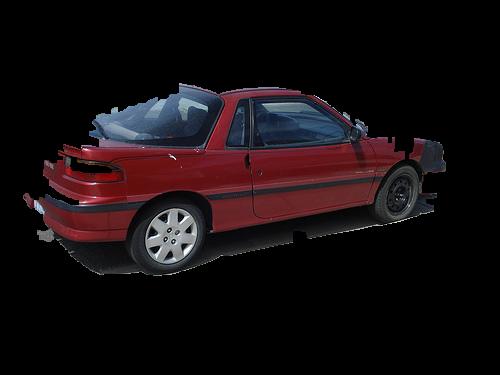}} &
\subfigure[]{\includegraphics[width=0.20\linewidth]{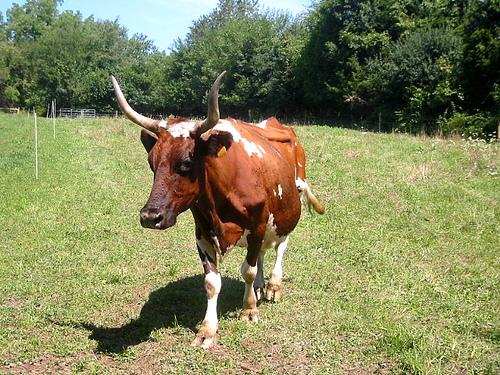}} &
\subfigure[]{\includegraphics[width=0.20\linewidth]{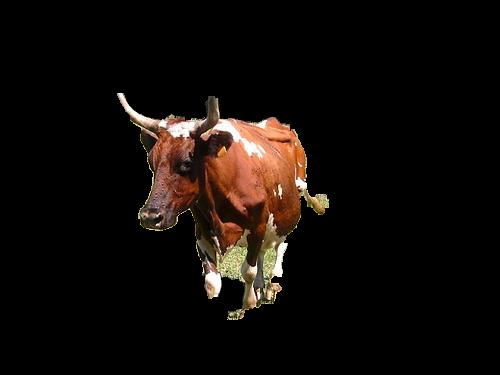}}\\
\subfigure[]{\includegraphics[width=0.20\linewidth]{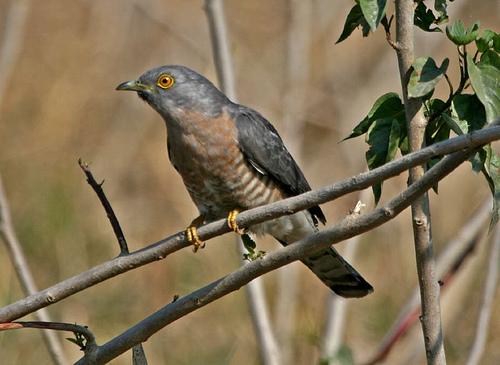}} &
\subfigure[]{\includegraphics[width=0.20\linewidth]{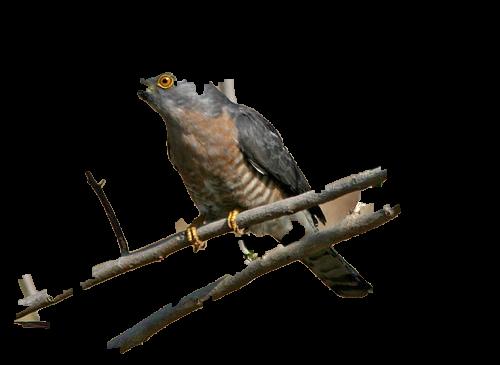}} &
\subfigure[]{\includegraphics[width=0.20\linewidth]{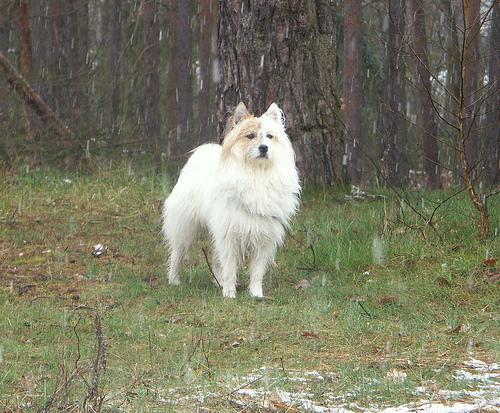}} &
\subfigure[]{\includegraphics[width=0.20\linewidth]{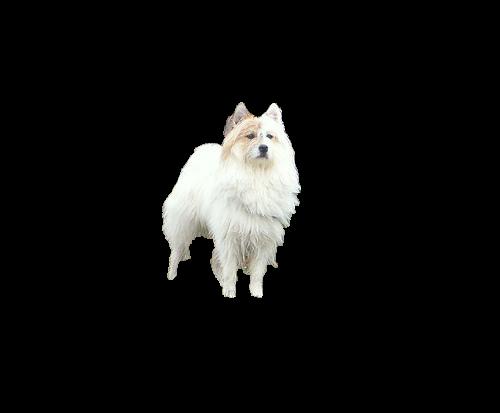}}
\end{tabular}
\caption{Representative segmentation results for the Pascal VOC 2011 dataset.
We model both the foreground and the background as GMMs with $K=3$ components.
The pairwise assignment probabilities utilize $m=3$ nearest neighbors, and the
prior by Rosenfeld et al. \cite{geometric} was used for initialization.}%
\label{fig:results_unsupervised}%
\end{figure}

Representative qualitative results are shown in Fig.
\ref{fig:results_unsupervised}.

It follows that the proposed scheme compares favorably with contemporary
state-of the-art schemes such as CPMC \cite{CPMC2} all using the same
initialization scheme, and performs similarly to Kuettel et al.
\cite{sliding_window}, whose initialization scheme is more elaborated.

\subsection{Runtime and Implementation}

The proposed scheme was implemented in unoptimized Matlab code, and the
simulations were ran on a Intel core i7 PC equipped with 8Gb of RAM. The
runtime of the algorithm is 1-2 seconds per iteration, such that the runtime
of 10 iterations is 15 seconds. As the distance matrix between the SPs is
computed once (in the first iteration), most of the running time is spent
computing the refined $\mathcal{F}$/$\mathcal{B}$ GMM models at each
iteration. The spectral solver detailed in Section \ref{subsec:assignment}
requires $\sim100ms$.

\section{Conclusions}

\label{sec:discussion}

In this work we presented an approach for Foreground/Backgrounds
($\mathcal{F}$/$\mathcal{B}$) image segmentation. Our scheme represents an
image as a set of SPs, that are classified to either $\mathcal{F}%
$/$\mathcal{B}$. The classification is formulated using a probabilistic
framework consisting of unary and pairwise probabilities estimated using
low-level image features. SPs are statistically characterized by Gaussian
models, allowing to define neighboring pairwise probabilities in terms of the
KL distances between neighboring superpixels. The foreground and background
are modeled via Gaussian mixture models, yielding a global representation of
both, and estimate unary assignment probability by way of KL distances between
the classes (GMMs) and the superpixels (Gaussians). The segmentation is thus
formulated as an inference problem over a two-dimensional MRF. For that we
applied a novel variation of a probabilistic inference scheme
\cite{KellerProbSpec}, which allows improved insights and an iterative
auto-tuning scheme for the parameters of the unary and pairwise assignment
probabilities. The proposed scheme is applied in both semi-supervised and
unsupervised settings, and is shown to compare favorably with state-of-the-art
schemes when applied to contemporary image testsets.

\FloatBarrier

\bibliographystyle{plain}
\bibliography{ImageSegmentationViaProbabilisticGraphMatching}

\begin{thebibliography}{10}

\bibitem{Alpert2012}
S.~Alpert, M.~Galun, A.~Brandt, and R.~Basri.
\newblock Image segmentation by probabilistic bottom-up aggregation and cue
  integration.
\newblock {\em {{IEEE} Transactions on Pattern Analysis and Machine
  Intelligence ({PAMI})}}, 34(2):315--327, 2012.

\bibitem{1238310}
Y.~Boykov and V.~Kolmogorov.
\newblock Computing geodesics and minimal surfaces via graph cuts.
\newblock In {\em {Proceedings of the International Conference on Computer
  Vision ({ICCV})}}, pages 26--33 vol.1, October 2003.

\bibitem{GraphCut}
Y.~Boykov and V.~Kolmogorov.
\newblock Computing geodesics and minimal surfaces via graph cuts.
\newblock In {\em {Proceedings of the International Conference on Computer
  Vision ({ICCV})}}, pages 26--33 vol.1, October 2003.

\bibitem{Boykov2001}
Yuri Boykov, Olga Veksler, and Ramin Zabih.
\newblock Fast approximate energy minimization via graph cuts.
\newblock {\em {{IEEE} Transactions on Pattern Analysis and Machine
  Intelligence ({PAMI})}}, 23(11):1222--1239, November 2001.

\bibitem{BoykovJolly}
Y.Y. Boykov and M.-P. Jolly.
\newblock Interactive graph cuts for optimal boundary amp; region segmentation
  of objects in n-d images.
\newblock In {\em {Proceedings of the International Conference on Computer
  Vision ({ICCV})}}, volume~1, pages 105--112 vol.1, 2001.

\bibitem{CPMC2}
J.~Carreira and C.~Sminchisescu.
\newblock Constrained parametric min-cuts for automatic object segmentation.
\newblock In {\em {Proceedings of the {IEEE} Conference on Computer Vision and
  Pattern Recognition ({CVPR})}}, pages 3241--3248, June 2010.

\bibitem{CPMC}
Joao Carreira and Cristian Sminchisescu.
\newblock {CPMC}: Automatic object segmentation using constrained parametric
  min-cuts.
\newblock {\em {{IEEE} Transactions on Pattern Analysis and Machine
  Intelligence ({PAMI})}}, 34(7):1312--1328, 2012.

\bibitem{Topological}
Chao Chen, D.~Freedman, and Christoph~H. Lampert.
\newblock Enforcing topological constraints in random field image segmentation.
\newblock In {\em {Proceedings of the {IEEE} Conference on Computer Vision and
  Pattern Recognition ({CVPR})}}, pages 2089--2096, June 2011.

\bibitem{CompactPrior}
P.~Das and O.~Veksler.
\newblock Semiautomatic segmentation with compact shapre prior.
\newblock In {\em Computer and Robot Vision, 2006. The 3rd Canadian Conference
  on}, page~28, June 2006.

\bibitem{KellerProbSpec}
A.~Egozi, Y.~Keller, and H.~Guterman.
\newblock A probabilistic approach to spectral graph matching.
\newblock {\em {{IEEE} Transactions on Pattern Analysis and Machine
  Intelligence ({PAMI})}}, 35(1):18--27, January 2013.

\bibitem{pascal-voc-2009}
M.~Everingham, L.~Van~Gool, C.~K.~I. Williams, J.~Winn, and A.~Zisserman.
\newblock The {PASCAL} {V}isual {O}bject {C}lasses {C}hallenge 2009 {(VOC2009)}
  {R}esults.
\newblock
  http://www.pascal-network.org/challenges/VOC/voc2009/workshop/index.html.

\bibitem{pascal-voc-2010}
M.~Everingham, L.~Van~Gool, C.~K.~I. Williams, J.~Winn, and A.~Zisserman.
\newblock The {PASCAL} {V}isual {O}bject {C}lasses {C}hallenge 2010 {(VOC2010)}
  {R}esults.
\newblock
  http://www.pascal-network.org/challenges/VOC/voc2010/workshop/index.html.

\bibitem{pascal-voc-2011}
M.~Everingham, L.~Van~Gool, C.~K.~I. Williams, J.~Winn, and A.~Zisserman.
\newblock The {PASCAL} {V}isual {O}bject {C}lasses {C}hallenge 2011 {(VOC2011)}
  {R}esults.
\newblock
  http://www.pascal-network.org/challenges/VOC/voc2011/workshop/index.html.

\bibitem{MedicalShapePrior}
D.~Freedman and Tao Zhang.
\newblock Interactive graph cut based segmentation with shape priors.
\newblock In {\em {Proceedings of the {IEEE} Conference on Computer Vision and
  Pattern Recognition ({CVPR})}}, volume~1, pages 755--762, June 2005.

\bibitem{girshick14CVPR}
R.~Girshick, J.~Donahue, T.~Darrell, and J.~Malik.
\newblock Rich feature hierarchies for accurate object detection and semantic
  segmentation.
\newblock In {\em {Proceedings of the {IEEE} Conference on Computer Vision and
  Pattern Recognition ({CVPR})}}, pages 580--587, Los Alamitos, CA, USA, June
  2014. IEEE Computer Society.

\bibitem{gold04}
J.~Goldberger, S.~Gordon, and H.~Greenspan.
\newblock An efficient image similarity measure based on approximations of
  kl-divergence between two gaussian mixtures.
\newblock In {\em {Proceedings of the International Conference on Computer
  Vision ({ICCV})}}, pages 487--493 vol.1, October 2003.

\bibitem{BharathECCV2014}
Bharath Hariharan, Pablo Arbel\'{a}ez, Ross Girshick, and Jitendra Malik.
\newblock Simultaneous detection and segmentation.
\newblock In {\em {Proceedings of the European Conference on Computer Vision
  ({ECCV})}}, 2014.

\bibitem{MalikShi}
S.~Jianbo and J.~Malik.
\newblock Normalized cuts and image segmentation.
\newblock {\em {{IEEE} Transactions on Pattern Analysis and Machine
  Intelligence ({PAMI})}}, 22(8):888 --905, August 2000.

\bibitem{NIPS2011_4296}
Philipp Kr\"{a}henb\"{u}hl and Vladlen Koltun.
\newblock Efficient inference in fully connected crfs with gaussian edge
  potentials.
\newblock In J.~Shawe-Taylor, R.S. Zemel, P.L. Bartlett, F.~Pereira, and K.Q.
  Weinberger, editors, {\em {Neural Information Processing Systems
  ({NeurIPS})}}, pages 109--117. Curran Associates, Inc., 2011.

\bibitem{sliding_window}
D.~Kuettel and V.~Ferrari.
\newblock Figure-ground segmentation by transferring window masks.
\newblock In {\em {Proceedings of the {IEEE} Conference on Computer Vision and
  Pattern Recognition ({CVPR})}}, pages 558--565, June 2012.

\bibitem{BoxPrior}
V.~Lempitsky, P.~Kohli, C.~Rother, and T.~Sharp.
\newblock Image segmentation with a bounding box prior.
\newblock In {\em {Proceedings of the International Conference on Computer
  Vision ({ICCV})}}, pages 277--284, September 2009.

\bibitem{Leordeanu-spectral}
M.~Leordeanu and M.~Hebert.
\newblock A spectral technique for correspondence problems using pairwise
  constraints.
\newblock In {\em {Proceedings of the International Conference on Computer
  Vision ({ICCV})}}, volume~2, pages 1482 -- 1489, October 2005.

\bibitem{turbo}
A.~Levinshtein, A.~Stere, K.N. Kutulakos, D.J. Fleet, S.J. Dickinson, and
  K.~Siddiqi.
\newblock {TurboPixels}: Fast superpixels using geometric flows.
\newblock {\em {{IEEE} Transactions on Pattern Analysis and Machine
  Intelligence ({PAMI})}}, 31(12):2290--2297, December 2009.

\bibitem{LazySnapping}
Yin Li, Jian Sun, Chi-Keung Tang, and Heung-Yeung Shum.
\newblock Lazy snapping.
\newblock {\em ACM Trans. Graph.}, 23(3):303--308, August 2004.

\bibitem{MemoryReduction}
H.~Lombaert, Yiyong Sun, L.~Grady, and Chenyang Xu.
\newblock A multilevel banded graph cuts method for fast image segmentation.
\newblock In {\em {Proceedings of the International Conference on Computer
  Vision ({ICCV})}}, volume~1, pages 259--265 Vol. 1, October 2005.

\bibitem{long_shelhamer_fcn}
Jonathan Long, Evan Shelhamer, and Trevor Darrell.
\newblock Fully convolutional networks for semantic segmentation.
\newblock {\em {Proceedings of the {IEEE} Conference on Computer Vision and
  Pattern Recognition ({CVPR})}}, November 2015.

\bibitem{MCLA2000}
G.~J. McLachlan and D.~Peel.
\newblock {\em Finite mixture models}.
\newblock Wiley Series in Probability and Statistics, New York, 2000.

\bibitem{NIPS2001_2092}
Andrew~Y. Ng, Michael~I. Jordan, and Yair Weiss.
\newblock On spectral clustering: Analysis and an algorithm.
\newblock In T.G. Dietterich, S.~Becker, and Z.~Ghahramani, editors, {\em
  {Neural Information Processing Systems ({NeurIPS})}}, pages 849--856. MIT
  Press, 2002.

\bibitem{Oliva}
Aude Oliva and Antonio Torralba.
\newblock Modeling the shape of the scene: A holistic representation of the
  spatial envelope.
\newblock {\em {International Journal of Computer Vision ({IJCV})}},
  42(3):145--175, May 2001.

\bibitem{GMM-BIC}
Ana Oliveira-Brochado and Francisco~Vitorino Martins.
\newblock {Assessing the Number of Components in Mixture Models: a Review}.
\newblock FEP Working Papers 194, Universidade do Porto, Faculdade de Economia
  do Porto, November 2005.

\bibitem{renNIPS15fasterrcnn}
Shaoqing Ren, Kaiming He, Ross Girshick, and Jian Sun.
\newblock Faster {R-CNN}: Towards real-time object detection with region
  proposal networks.
\newblock In {\em {Neural Information Processing Systems ({NeurIPS})}}, 2015.

\bibitem{geometric}
A.~Rosenfeld and D.~Weinshall.
\newblock Extracting foreground masks towards object recognition.
\newblock In {\em {Proceedings of the International Conference on Computer
  Vision ({ICCV})}}, pages 1371 --1378, nov. 2011.

\bibitem{RotherKB}
Carsten Rother, Vladimir Kolmogorov, and Andrew Blake.
\newblock "grabcut": Interactive foreground extraction using iterated graph
  cuts.
\newblock {\em ACM Trans. Graph.}, 23(3):309--314, August 2004.

\bibitem{KellerOfdm}
A.~Septimus, Y.~Keller, and I.~Bergel.
\newblock A spectral approach to inter-carrier interference mitigation in ofdm
  systems.
\newblock {\em Communications, IEEE Transactions on}, 62(8):2802--2811, August
  2014.

\bibitem{ShiNCI}
Jianbo Shi and Jitendra Malik.
\newblock Normalized cuts and image segmentation.
\newblock {\em {{IEEE} Transactions on Pattern Analysis and Machine
  Intelligence ({PAMI})}}, 22(8):888--905, August 2000.

\bibitem{EllipticalShapePrior}
G.~Slabaugh and G.~Unal.
\newblock Graph cuts segmentation using an elliptical shape prior.
\newblock In {\em {Proceedings of the {IEEE} International Conference on Image
  Processing ({ICIP})}}, volume~2, pages II--1222, 2005.

\bibitem{5204091}
K.E.A. van~de Sande, T.~Gevers, and C.G.M. Snoek.
\newblock Evaluating color descriptors for object and scene recognition.
\newblock {\em {{IEEE} Transactions on Pattern Analysis and Machine
  Intelligence ({PAMI})}}, 32(9):1582--1596, September 2010.

\bibitem{starPrior}
Olga Veksler.
\newblock Star shape prior for graph-cut image segmentation.
\newblock In {\em Proceedings of the 10th European Conference on Computer
  Vision: Part III}, {Proceedings of the European Conference on Computer Vision
  ({ECCV})}, pages 454--467, Berlin, Heidelberg, 2008. Springer-Verlag.

\bibitem{WatershedB}
L.~Vincent and P.~Soille.
\newblock Watersheds in digital spaces: An efficient algorithm based on
  immersion simulations.
\newblock {\em {{IEEE} Transactions on Pattern Analysis and Machine
  Intelligence ({PAMI})}}, 13:583--598, 1991.

\bibitem{Weiss2001}
Y.~Weiss and W.~T. Freeman.
\newblock On the optimality of solutions of the max-product belief-propagation
  algorithm in arbitrary graphs.
\newblock {\em {{IEEE} Transactions on Information Theory}}, 47(2):736--744,
  September 2001.

\bibitem{GrabcutCode}
Ming Xiumingzhang.
\newblock {GrabCut Matlab} implmentation.
\newblock {https://github.com/xiumingzhang/grabcut}.

\bibitem{Xu}
Linli Xu, Wenye Li, and D.~Schuurmans.
\newblock Fast normalized cut with linear constraints.
\newblock In {\em {Proceedings of the {IEEE} Conference on Computer Vision and
  Pattern Recognition ({CVPR})}}, pages 2866--2873, June 2009.

\bibitem{5539797}
Qingxiong Yang, Liang Wang, and N.~Ahuja.
\newblock A constant-space belief propagation algorithm for stereo matching.
\newblock In {\em {Proceedings of the {IEEE} Conference on Computer Vision and
  Pattern Recognition ({CVPR})}}, pages 1458--1465, June 2010.

\bibitem{Yu}
S.X. Yu and Jianbo Shi.
\newblock Segmentation given partial grouping constraints.
\newblock {\em {{IEEE} Transactions on Pattern Analysis and Machine
  Intelligence ({PAMI})}}, 26(2):173--183, Feb 2004.

\bibitem{Zhen_ICCV15_CRFRNN}
S.~Zheng, S.~Jayasumana, B.~Romera-Paredes, V.~Vineet, Z.~Su, D.~Du, C.~Huang,
  and P.H.S. Torr.
\newblock {Conditional Random Fields as Recurrent Neural Networks}.
\newblock In {\em {Proceedings of the International Conference on Computer
  Vision ({ICCV})}}, 2015.

\end{thebibliography}

\begin{center}
{\includegraphics[width=0.40\linewidth]{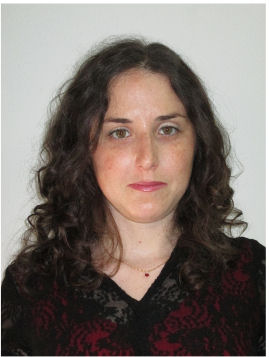} }
\end{center}
\textbf{Ayelet Heimowitz} received the BSc degree in Computer Engineering in 2009 from Bar Ilan University,
Israel. She received the MSc degree in Electrical Engineering in 2011. She is currently studying toward the Ph.D. degree in Electrical Engineering in the Faculty of Engineering, Bar-Ilan University, Ramat-Gan, Israel.

\begin{center}
\includegraphics[width=0.40\linewidth]{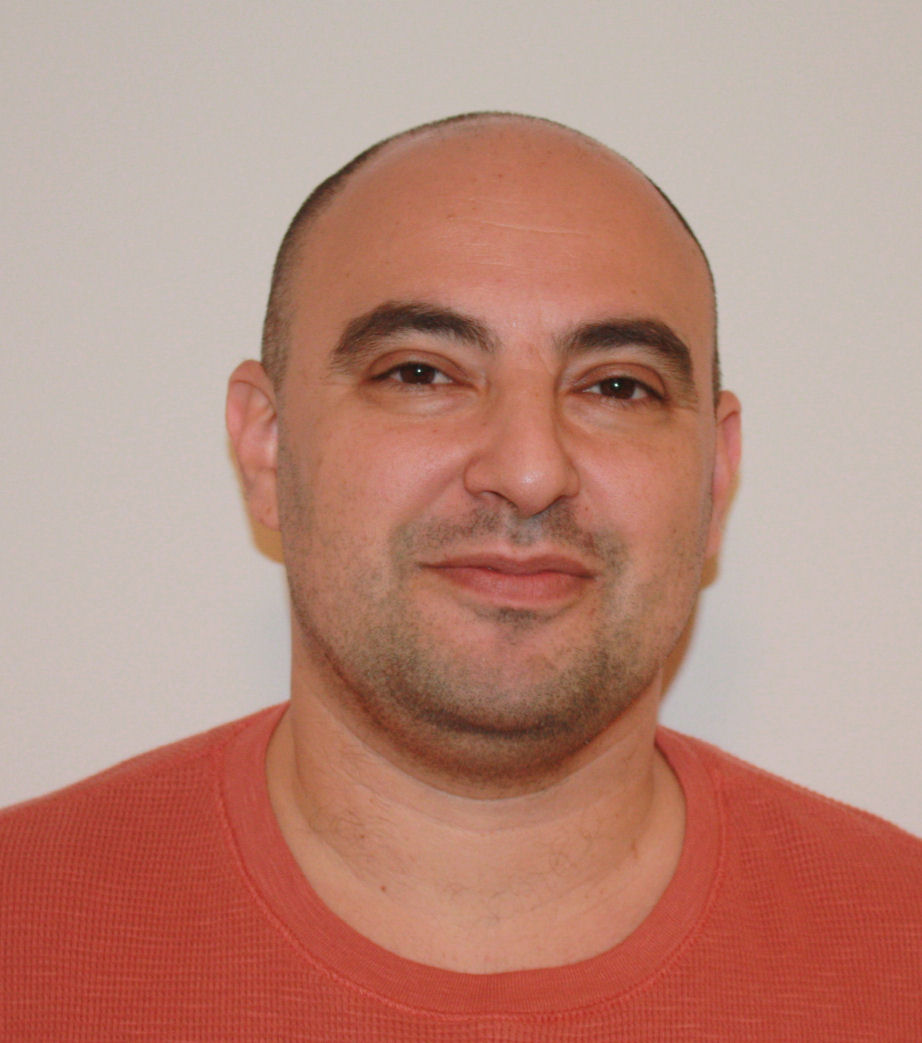}
\end{center}
\textbf{Yosi Keller} received the BSc degree in Electrical Engineering in 1994 from the Technion-Israel
Institute of Technology, Haifa. He received the MSc and PhD degrees in electrical engineering
from Tel-Aviv University, Tel-Aviv, in 1998 and 2003, respectively. From 2003 to 2006 he was a
Gibbs assistant professor with the Department of Mathematics, Yale University. He is an Associate
Professor at the Faculty of Engineering in Bar Ilan University, Israel.

\end{document}